\documentclass[10pt,twocolumn,letterpaper]{article}

\usepackage{cvpr}
\usepackage{times}
\usepackage{epsfig}
\usepackage{graphicx}
\usepackage{amsmath}
\usepackage{amsthm}
\usepackage{amssymb}
\usepackage{subfig}
\usepackage{verbatim}
\usepackage{booktabs}

\graphicspath{{Images/}}

\usepackage[pagebackref=true,breaklinks=true,letterpaper=true,colorlinks,bookmarks=false]{hyperref}

\def\cvprPaperID{3112}

\usepackage[algo2e,ruled,vlined]{algorithm2e}

\renewcommand{\vec}[1]{\mathbf{#1}}

\newcommand{\mr}[1]{\mathrm{#1}}
\newcommand{\tx}[1]{\textrm{#1}}
\newcommand{\tr}[1]{{#1}^\top}

\newcommand{\Om}{\Omega}

\newcommand{\RR}{\mathbb{R}}

\newcommand{\NN}{\mathcal{N}}

\newcommand{\xx}{\vec x}

\newcommand{\va}{\vec a}

\newcommand{\hy}{\widehat{\vec y}}
\newcommand{\hu}{\widehat{\vec u}}
\newcommand{\hW}{\widehat{W}}
\newcommand{\hD}{\widehat{D}}
\newcommand{\hL}{\widehat{L}}

\newcommand{\yy}{\vec y}
\newcommand{\vo}{\vec 1}
\newcommand{\vz}{\vec 0}

\newcommand{\uu}{\vec u}

\DeclareMathOperator*{\argmin}{arg\,min}

\newcommand{\pr}[2]{p(#1 \, | \, #2)}

\newcommand{\beq}{\begin{equation}}
\newcommand{\eeq}{\end{equation}}
\newcommand{\beqa}{\begin{eqnarray}}
\newcommand{\eeqa}{\end{eqnarray}}
\newcommand{\beqas}{\begin{eqnarray*}}
\newcommand{\eeqas}{\end{eqnarray*}}

\newcommand{\nnb}{\nonumber}

\newcommand{\dope}{\textsc{Dope}}

\usepackage{color}
\newcounter{cntcomment}

\renewcommand{\paragraph}[1]{\vspace{0.5em}\noindent{\textbf{{#1}}}~}

\expandafter\def\expandafter\normalsize\expandafter{%
    \normalsize
    \setlength\abovedisplayskip{3pt}
    \setlength\belowdisplayshortskip{3pt}
}
\newtheorem{theorem1}{Theorem}

\newtheorem{proposition1}{Proposition}

\cvprfinalcopy 

\begin{document}

\title{
DOPE: Distributed Optimization for Pairwise Energies 
}

\author{Jose Dolz\\
{\tt\small jose.dolz@livia.etsmtl.ca}
\and
Ismail Ben Ayed\\
{\tt\small ismail.benayed@etsmtl.ca}
\and
Christian Desrosiers\\
Laboratory for Imagery, Vision and Artificial Intelligence\\
Ecole de Technologie Superieure, Montreal, Canada\\
{\tt\small christian.desrosiers@etsmtl.ca}
}

\maketitle

\begin{abstract}
We formulate an Alternating Direction Method of Multipliers (ADMM) that systematically distributes the computations of any technique for optimizing pairwise functions, including  non-submodular potentials. Such discrete functions are very useful in segmentation and a breadth of other vision problems. Our method decomposes the problem into a large set of small sub-problems, each involving a sub-region of the image domain, which can be solved in parallel. We achieve consistency between the sub-problems through a novel constraint that can be used for a large class of pairwise functions. We give an iterative numerical solution that alternates between solving the sub-problems and updating consistency variables, until convergence. We report comprehensive experiments, which demonstrate the benefit of our general distributed solution in the case of the popular serial algorithm of Boykov and Kolmogorov (BK algorithm) and, also, in the context of non-submodular functions. 

\end{abstract}

\section{Introduction}\label{sec:intro}

A mainstay in computer vision, regularization serves a breadth of applications and problems including segmentation \cite{Tang2015,Taniai2015}, optical flow \cite{Lempitsky2008}, shape fitting \cite{Lempitsky2007}, stereo matching \cite{Kolmogorov2001}, deconvolution \cite{Gorelick2014}, high-dimensional clustering \cite{Tang2016}, among many others \cite{Blake2011, Kappes2015}. For instance, in the discrete setting, segmentation problems are commonly stated as optimizing a regularization-based functional\footnote{We give a binary
(two-region) segmentation functional for simplicity but the discussion extends to multi-region segmentation.} of the following general form 
\cite{Baque2016,Boykov2006,Krahenbuhl2011}:
\begin{equation}
\label{General_Unary_Pairwise_form}
	E(\yy) \ = \ \sum_{i \in \Omega} u_i \, y_i  \ + \ \lambda \!\!\sum_{i,j \in \Om^2} \!\! w_{ij} \,  | y_i - y_j | 
\end{equation}
where $\Om$ is the image domain and  $\yy = \tr{(y_1, y_2, \dots, y_n)} \in \{0,1\}^{|\Om|}$ is a binary vector indicating a possible foreground-background segmentation: $y_i = 1$ if pixel $i$ belongs to the foreground class, otherwise  $y_i = 0$. $\lambda \geq 0$ controls the relative importance of each term. The first term is a sum of unary potentials typically defined via log posteriors:
\beq
	u_i \ = \ \log \pr{y_i = 0}{{\mathbf x}_i} - \log \pr{y_i = 1}{{\mathbf x}_i}
\eeq
with ${\mathbf x}_i \in \RR^M$ denoting the feature vector of pixel $i \in \Om$ (e.g., color). The second term in \eqref{General_Unary_Pairwise_form} is a general form of {\em pairwise} regularization. The second-order Potts model \cite{Boykov2006} is an important example of pairwise regularization, and is very popular in computer vision: given a neighborhood $\NN(i)$ for pixel $i$, 
$w_{ij} >0 \mbox{~if~} j \in \NN(i)$ and $0$ elsewhere.
In this case, $w_{ij}$ is a penalty for assigning different labels to neighboring pixels $i$ and $j$. Such a penalty can be either a constant, in which case the regularization term measures 
the length of segment boundary, or a decreasing function of feature (e.g., color) difference $\|\xx_i - \xx_j \|$, which attracts the segment boundary towards strong feature edges \cite{Boykov2006}. 
Potts regularization belongs to an important family of discrete pairwise functions, {\em submodular functions}\footnote{A function $f$ defined over a pair of discrete binary variables is submodular if and only if $f(1,0) + f(0,1) \geq f(1,1) + f(0,0)$.}, which were instrumental in the development of various efficient computer vision algorithms. The global optimum of a function containing unary and submodular pairwise potentials can be computed exactly in polynomial time using graph cut (or max-flow) algorithms \cite{Boykov2004}. Other examples of pairwise terms of the general form in \eqref{General_Unary_Pairwise_form} include non-submodular functions, which arise in problems such as curvature regularization \cite{Gorelick2014,Nieuwenhuis2014}, surface registration \cite{Kappes2015}, deconvolution \cite{Gorelick2014} and inpainting \cite{Kappes2015}. It also includes dense (fully connected) models \cite{Krahenbuhl2011}, where pairwise penalties $w_{ij}$ are not restricted to neighbouring pixels. These are only few examples of pairwise-function problems widely used in combination with popular optimization techniques such as LP relaxation \cite{Kappes2015} or mean-field inference \cite{Baque2016,Krahenbuhl2011}. 
Finally, it is worth mentioning that total-variation (TV) terms can be viewed as the continuous counterpart of Potts regularization, and were the subject of a large number of vision works in recent years \cite{Chambolle2011,Punithakumar2012,Souiai2015,Yuan:CVPR2010}.

Recently, there have been significant research efforts focusing on designing parallel (or distributed) formulations for optimizing pairwise functions \cite{Baque2016,Liu2010,Strandmark2010}. Distributing computations would be beneficial not only to high-resolution images and massive 3D grids but also to difficult high-order models \cite{Gorelick2013,Kee2014,Tang2014}, which require approximate solutions solving a large number of problems of the form \eqref{General_Unary_Pairwise_form}. For instance, continuous convex relaxation techniques have been gaining popularity recently due to their ability to accommodate parallel implementations \cite{Chambolle2011,Punithakumar2012,Souiai2015,Yuan:CVPR2010}. Unfortunately such techniques are restricted to TV regularization terms. Mean-field inference techniques \cite{Baque2016,Krahenbuhl2011} also attracted significant attention recently as they can be parrallelized, albeit at the cost of convergence guarantees \cite{Baque2016}.

In the context of submodular functions, several studies focused specifically on parallel formulations of the max-flow/graph-cut algorithm of Boykov and Kolmogorov (BK algorithm) \cite{Boykov2004}, which has made a substantial impact in computer vision. The BK algorithm yields a state-of-the-art empirical performance for typical vision problems such as 2D segmentation. Even though this augmenting path algorithm is serial, it uses heuristics that handle efficiently sparse 2D grids, outperforming top parallel push-relabel max-flow algorithms \cite{Delong2008}. Unfortunately, distributing the computations for the BK algorithm is not a trivial problem\footnote{Augmenting-path max-flow algorithms are based on global operations and, therefore, do not accommodate parallel/distributed implementations}, and the efficiency of the algorithm may decrease when moving from 2D to 3D (or higher-dimensional) grids. Therefore, parallel/distributed computations for this algorithm would be of substantial benefit to the community, and several works addressed the problem \cite{Bhusnurmath2008,Liu2010,Strandmark2010}. For instance, the method in \cite{Liu2010} investigated a bottom-up approach to parallelize the BK algorithm using two subsequent phases: the first stage partitions the graph into several sub-graphs and processes them in parallel, whereas the second stage gradually merges the subgraphs so as to involve longer paths, until a global minimum is reached. Unfortunately, this technique requires a shared-memory model, which does not accommodate distributed computations. The method in \cite{Strandmark2010} wrote the max-flow (graph cut) problem as a linear program, and viewed the objective function as a sum of two functions, each involving a sub-graph. Then, they used a dual decomposition formulation to process each of the two sub-graphs independently. However, it is not clear how to split the problem into a large number of sub-graphs (for faster computations) as this would increase exponentially the number of constraints (w.r.t. the sub-graphs). 
The method in \cite{Bhusnurmath2008} proposed a linear program formulation of the BK algorithm, via a $L_1$ minimization statement. Solving the problem via Newton iterations yields matrix-vector multiplications, which can be evaluated in parallel. This method, however, is not significantly faster than the serial BK algorithm \cite{Strandmark2010}.  

In general, the existing distributed/parallel formulations for optimizing pairwise functions are technique-specific. For instance, the methods in \cite{Bhusnurmath2008,Liu2010,Strandmark2010} were tailored for the BK algorithm,
and it is not clear how to extend these methods beyond the context of max-flow formulations and submodular functions. In this study, we formulate an Alternating Direction Method of Multipliers (ADMM), which systematically distributes the computations of any technique for optimizing pairwise functions, including  non-submodular potentials. Our method decomposes the problem into a large set of small sub-problems, each involving a sub-region of the image domain (i.e., block), which can be solved in parallel. We achieve consistency between the sub-problems through a novel constraint that can be used in conjunction with any functional
of the form \eqref{General_Unary_Pairwise_form}. We give an iterative numerical solution that alternates between solving the sub-problems and updating consistency variables, until convergence. Our method can be viewed as a variant of the alternating projections algorithm to find a point in the intersection of two convex sets and, therefore, is well suited to distributed convex optimization. We report comprehensive experiments, which demonstrate the benefit of our general solution in the case of the popular BK algorithm and, also, in the context of non-submodular functions.

\section{Formulation}

Let $\uu =  \tr{(u_1, u_2, \dots, u_n)} \in \RR^{|\Om|}$ be the vector of unary penalties and $W \in \RR^{|\Om| \times |\Om|}$ the matrix of pairwise penalties $w_{ij}$. It is easy to show that the general segmentation problem in \eqref{General_Unary_Pairwise_form} can be expressed in matrix form as follows:
\beq\label{eq:cost-function}
	\argmin_{\yy \, \in \, \{0,1\}^{|\Om|}} \ \tr{\uu} \yy \ + \ \lambda \tr{\yy} L \yy.
\eeq 
Here, $L=D-W$ is the Laplacian matrix corresponding to $W$, and $D$ is a diagonal matrix such that $d_{ii} = \sum_j w_{ij}$.

Let us divide a large image $\Om$ into $K$ blocks $\Om_k$ ($k=1,\ldots,K$) that can overlap, which allows pixels of image $\Om$ to be simultaneously located in multiple blocks. Let $\hy_k \in \{0,1\}^{|\Om_k|}$ denote the segmentation vector of block $k$. Our goal is to reformulate problem (\ref{eq:cost-function}) in a way that the tasks of segmenting blocks are not directly coupled, thus allowing them to be performed simultaneously. To achieve this, we connect them through the segmentation vector $\yy \in \{0,1\}^{|\Om|}$ of the whole image $\Om$, by imposing linear constraints $\hy_k = S_k \yy$, $k=1,\ldots, K$, where $S_k$ is a $|\Om_k| \times |\Om|$ matrix selecting the pixels of block $k$. 

Given the segmentation vectors of each block, global segmentation $\yy$ can be expressed using the following proposition.
\begin{proposition1}
If each pixel of $\Om$ belongs to at least one block, i.e. $\cup_{k=1}^K \Om_k = \Om$, and $\hy_k = S_k \yy$, $k=1,\ldots, K$, then the following relationship holds:
\beq\label{th:prop1}
	\yy \ = \ \Big( \sum_{k=1}^K \tr{S}_k S_k\Big)^{-1} \Big(\sum_{k=1}^K \tr{S}_k \hy_k \Big).
\eeq
\end{proposition1}
Here, $Q = \sum_k \tr{S}_k S_k$ is a diagonal matrix such that $q_{ii}$ is the number of blocks containing pixel $i$. In short, this proposition states that, if the block segmentation vectors are consistent (i.e., pixels have the same label across blocks containing them), then $y_i$ is simply the mean label of pixel $i$ within the blocks containing this pixel. This property also applies when relaxing the integer constraints on $\yy$, allowing us to develop an efficient optimization strategy.

In the following theorem, we show that segmentation problem (\ref{eq:cost-function}) can be reformulated as a sum of similar sub-problems, one for each block $k$, connected together through a relaxed global segmentation vector $\yy$.
\begin{theorem1}\label{th:theo1}
Denote as $\hu_k = S_k Q^{-1} \uu$ and $\hW_k = S_k Q^{-1} W Q^{-1} \tr{S}_k$ the unary and binary potential weights of block $k$, adjusted to consider the occurrence of pixels in multiple blocks. Moreover, let $\hD_k$ be the diagonal matrix such that  $[\hD_k]_{ii} = \sum_j [\hW_k]_{ij}$, and $\hL_k = \hD_k - \hW_k$ be the Laplacian of $\hW_k$. If $\hy_k = S_k \yy, \ k=1,\ldots, K$, then problem (\ref{eq:cost-function}) can be reformulated as
\begin{align}\label{eq:final-formulation}
	\argmin_{\substack{\yy \, \in \, \RR^{|\Om|} \\ \hy_k \, \in \, \{0,1\}^{|\Om_k|} }} & \  
		\sum_{k=1}^K \tr{\big(\hu_k \, + \, \lambda (C_k + R_k S_k) \yy\big)} \hy_k \nnb\\[-5mm]
		 & + \lambda \sum_{k=1}^K \tr{\hy}_k \hL_k \hy_k
\end{align}
where 
$C_k = S_k Q^{-1} L (I - Q^{-1} \tr{S}_k S_k)$ and $R_k = S_k Q^{-1} D Q^{-1} \tr{S}_k \, - \, \hD_k$.
\end{theorem1}
\begin{proof}
See details in Appendix \ref{sec:proof}.
\end{proof}

We solve problem (\ref{eq:final-formulation}) with an ADMM approach. Moving constraints $\hy_k = S_k \yy, \ k=1,\ldots, K$, into the functional via augmented Lagrangian terms \cite{hestenes1969multiplier} (with multiplier $\va_k$) gives:
\begin{multline}\label{eq:admm-formulation}
\argmin_{
	\substack{\yy \, \in \, \RR^{|\Om|}, \, \hy_k \in  \{0,1\}^{|\Om_k|} 
					\\ \va_k \, \in \, \RR^{|\Om_k|} }} \ \sum_{k=1}^K \tr{\big(\hu_k + \lambda (C_k + R_k S_k) \yy\big)} \hy_k \\[-1mm]
					 \ \ \ + \lambda \sum_{k=1}^K \tr{\hy}_k \hL_k \hy_k 
					+ \frac{\mu}{2} \sum_{k=1}^K \| \hy_k - S_k \yy + \va_k\|_2^2.
\end{multline}     		
     		     		
In this equation, augmented Lagrangian parameter $\mu \geq 0$ controls the trade-off between the original functional and satisfying the constraints. In general, ADMM methods are not overly sensitive to this parameter and converge if $\mu$ is large enough \cite{bras2012alternating}. In practice, $\mu$ is initialized using a small value and increased at each iteration by a given factor. To solve problem (\ref{eq:admm-formulation}), we note that the functional is convex with respect to each parameter $\hy_k$, $\yy$ and $\va_k$. We thus update these parameters alternatively, until convergence is reached (i.e., the constraints are satisfied up to a given $\epsilon$). 


Given $\yy$, the segmentation vectors of each block $k$ can be updated independently, in parallel, by solving the following problem:
 \begin{multline}
\argmin_{\hy_k \, \in \, \{0,1\}^{|\Om_k|} }   \	\ \tr{\big(\hu_k \, + \, \lambda (C_k + R_k S_k) \yy\big)} \hy_k 
     		\ + \ \lambda \tr{\hy}_k \hL_k \hy_k \\[-1mm]
     		 \ \  + \ \frac{\mu}{2} \| \hy_k \, - \, (S_k \yy - \va_k)\|_2^2
\end{multline}
Using the fact that $\|\hy_k\|^2_2 = \tr{\vo}\hy_k$ for binary vector $\hy_k$, we reformulate the problem as:
 \begin{multline}\label{eq:admm-updateYk}
\argmin_{\hy_k \, \in \, \{0,1\}^{|\Om_k|} }  \Big(\hu_k \, + \, \lambda \big(C_k + R_k S_k\big) \yy  \\[-1mm]
	+  \ \mu \big(\va_k - S_k \yy + \tfrac{1}{2}\big) \Big)\tr{} \hy_k  \ + \ \lambda \tr{\hy}_k \hL_k \hy_k.
\end{multline}
Notice that for $\yy$ fixed, this block problem corresponds
to a sum of unary and pairwise potentials. Therefore, as discussed in the introduction, it can be solved with one of the popular techniques\footnote{The choice depends on the form of the matrix of pairwise potentials.}, e.g., the BK algorithm \cite{Boykov2004} 

Once all block segmentation vectors have been computed, we can update the global segmentation $\yy$ by solving the following problem:
\begin{multline}\label{eq:admm-updateY}
	\argmin_{\yy \, \in \, \RR^{|\Om|}}  \ \ \lambda \sum_{k=1}^K  \tr{\hy}_k (C_k + R_k S_k) \yy \\[-2mm]
     	 + \frac{\mu}{2} \sum_{k=1}^K \|  S_k \yy  \, - \, (\hy_k + \va_k)\|_2^2.
\end{multline}
Since we have relaxed the integer constraints on $\yy$, this corresponds to a unconstrained least-square problem, whose solution is given by:
\beq\label{eq:admm-updateY2}
	\yy \ = \ \frac{1}{\mu} Q^{-1} \sum_{k=1}^K \Big( \mu \tr{S}_k (\hy_k + \va_k) - \lambda \tr{(C_k + R_k S_k)} \hy_k \Big).
\eeq
Note that since $Q$ is diagonal, computing its inverse is trivial.

Finally, the Lagrangian multipliers are updated as in standard ADMM methods:
\beq\label{eq:admm-updateAk}
	\va_k \ = \ \va_k \, + \, (\hy_k - S_k \yy), \ \ k=1,\ldots,K.
\eeq

The pseudo-code for implementing our \dope{} method is given in Algorithm \ref{alg:the_alg}. In a first step, the algorithm computes the unary and pairwise potentials of the global image, and divides the image into possibly overlapping blocks $\Om_k$, $k=1,\ldots,K$, based on a given partition scheme. For each block $k$, the algorithm pre-computes parameters $S_k$, $C_k$, $R_k$, $\hW_k$ and $\hu_k$. Note that these parameters can be computed in parallel. In the main loop, the algorithm then simultaneously recomputes segmentation vectors $\hy_k$ of each block, and uses them to update the global segmentation vector $\yy$. This process is repeated until constraints linking the block segmentation to the global segmentation are satisfied, up to a given $\epsilon \geq 0$. As mentioned earlier, the algorithm's convergence is facilitated by increasing the ADMM parameter $\mu$ by a factor of $\mu_\mr{fact}$ at each iteration.


\begin{algorithm2e}[tbh!]
\small
    \KwIn{The input image $\Om$ and pixel features $\xx$;}
    \KwIn{Block partition scheme;}
    \KwIn{Parameters $\lambda$, $\epsilon$, $\mu_0$, $\mu_\mr{fact}$;}
    \KwOut{The segmentation vector $\yy$;}
    \caption{\mbox{\dope{} segmentation algorithm}}  
    \label{alg:the_alg}  
    
    \BlankLine 
    \emph{Initialization}: \\    
    Compute global image unary and pairwise potentials\; 
  	Compute blocks and their corresponding parameters\;
    Set $y_i : = \tfrac{1}{2}$, $i=1,\ldots, |\Om|$\;    
    Set $\va_k := \vz$, $\ k=1,\ldots K$\;
    Set $\mu := \mu_0$\;
    	      	   
    \BlankLine
    \emph{Main loop}: \\  
	\While{$\exists k, \tx{ s.t. } \|\hy_k - S_k\yy\|_2 > \epsilon$}{	   
		\textbf{In parallel}: update $\hy_k$, $k=1,\ldots,K$, by solving (\ref{eq:admm-updateYk})\;
		Update $\yy$ by Eq. (\ref{eq:admm-updateY2})\;
	    Update $\va_k$, $k=1,\ldots,K$, based on Eq. (\ref{eq:admm-updateAk})\;
	    Set $\mu := \mu \times \mu_\mr{fact}$\;
	  }
	 	 
    \BlankLine
    \Return{$\yy$.}
\end{algorithm2e}



%



\section{Experiments}

The main goal of our experiments is to demonstrate that our \dope{} formulation can distribute the computations of powerful 
serial algorithms without affecting the quality of the energies at convergence. First, we prove that our formulation can achieve segmentation results consistent with the popular serial graph cut (sGC) algorithm of Boykov-Kolmogorov \cite{Boykov2004}, while allowing distributed computations. We illustrate the usefulness of our method on the task of segmenting high-resolution 2D multi-channel images (Section \ref{ssection:2D}) and 3D MRI brain volumes (Section \ref{ssection:3D}) using the second-order Potts model, and compare its accuracy, obtained energies and efficiency to those corresponding to sGC. The consistency between our method's segmentation and sGC is measured using Dice score coefficient (DSC) and relative energy differences: 
\beq\label{VolDif}
 \Delta E(\%) = \frac{ E_{GC} - E_{distReg}}{E_{GC}} \times 100
\eeq
where $E_{GC}$ is the energy of serial GC and $E_{distReg}$ is the energy given by our distributed regularization formulation.

Another objective of these experiments is to assess the impact of our method's parameters on computation time and segmentation accuracy. In particular, we evaluated how the partitioning scheme (i.e., block size and overlap) affects the method's performance. If blocks are small, a greater level of parallelism can be achieved, but segmentation consistency across blocks might be harder to satisfy. Conversely, using larger blocks with more overlap encourages global consistency of the segmentation, but might increase the run times. In our experiments, we considered three partitioning schemes, dividing images into $K= 32$, $64$ or $128$ even-sized blocks. For each of these, we tested three levels of overlap. In the first one, denoted by $\mr{Size}_{00}$, images were split into $K$ non-overlapping blocks covering the whole image (i.e., each pixel/voxel is in exactly one block). The size of these blocks was then increased by 10\% and 25\%, leading to larger blocks with greater overlap. We denote these two overlapping partitions by $\mr{Size}_{10}$ and $\mr{Size}_{25}$, respectively. Furthermore, we investigated the impact of neighbourhood size (i.e., the number of non-zero pairwise potentials $w_{ij}$) on segmentation performance. Using larger neighbourhoods, as defined by the kernel, can lead to a finer segmentation but significantly increases  run times. Kernel sizes of 3, 5, 7 and 9 pixels/voxels were considered in our experiments. We used square kernels for 2D images, and spherical kernels in the 3D setting. The regularization parameter $\lambda$ was selected per image (typically, its values are proportional to image size). Note that the \emph{same} $\lambda$ was used for computing the energy of both our method and sGC. Finally, the ADMM parameter was initialized to $\mu_0 = 100$ for 2D images and $\mu_0 = 500$ for 3D volumes, and increased by a factor of $\mu_\mr{fact} = 1.05$ at each iteration. 

Finally, we report curvature regularization experiments to show the use of our formulation in the case of non-submodular pairwise functions (Section \ref{ssection:curv}). These experiments involve distributing the computations of the trust region (LSA-TR) method in \cite{Gorelick2014}, a serial non-submodular optimizer that recently obtained competitive\footnote{LSA-TR outperforms significantly popular non-submodular optimization techniques such as TRWS and QPBO; See the comparative energy plots in \cite{Gorelick2014}.} performances in a wide range of applications (deconvolution, inpainting, among others). This shows that our formulation can be readily used in these applications.


Our method was implemented in \textsc{Matlab} R2015b, and all experiments were performed on a server with the following hardware specifications: 64 Intel(R) Xeon(R) 2.30GHz CPUs with 8 cores, and 128 GB RAM. For sGC, we used the publicly available B-K \textsc{Matlab} tool\footnote{\url{http://vision.csd.uwo.ca/code/}}, which implements the max-flow algorithm. In the next sections, we present the results obtained for high resolution 2D multi-channel images, 3D MRI data and a squared curvature regularization example.

\subsection{High-resolution 2D multi-channel images}
\label{ssection:2D}

\begin{figure*}[htb!]
     \begin{center}
        \mbox{
        \shortstack{
             \includegraphics[width=0.175\linewidth]{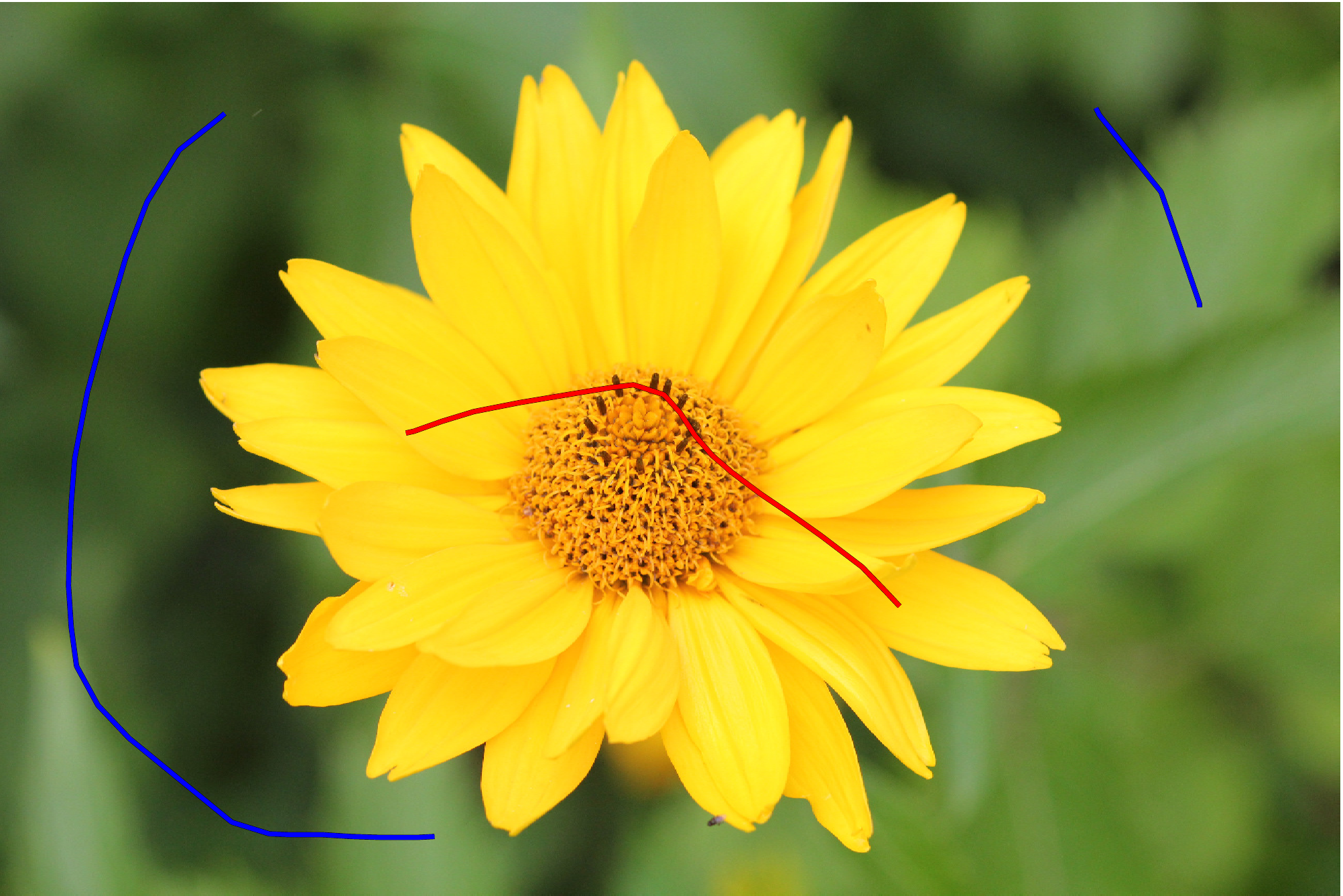}\\
            a) Original image}
        \hspace{-0.25em}
        \shortstack{
             \includegraphics[width=0.175\linewidth]{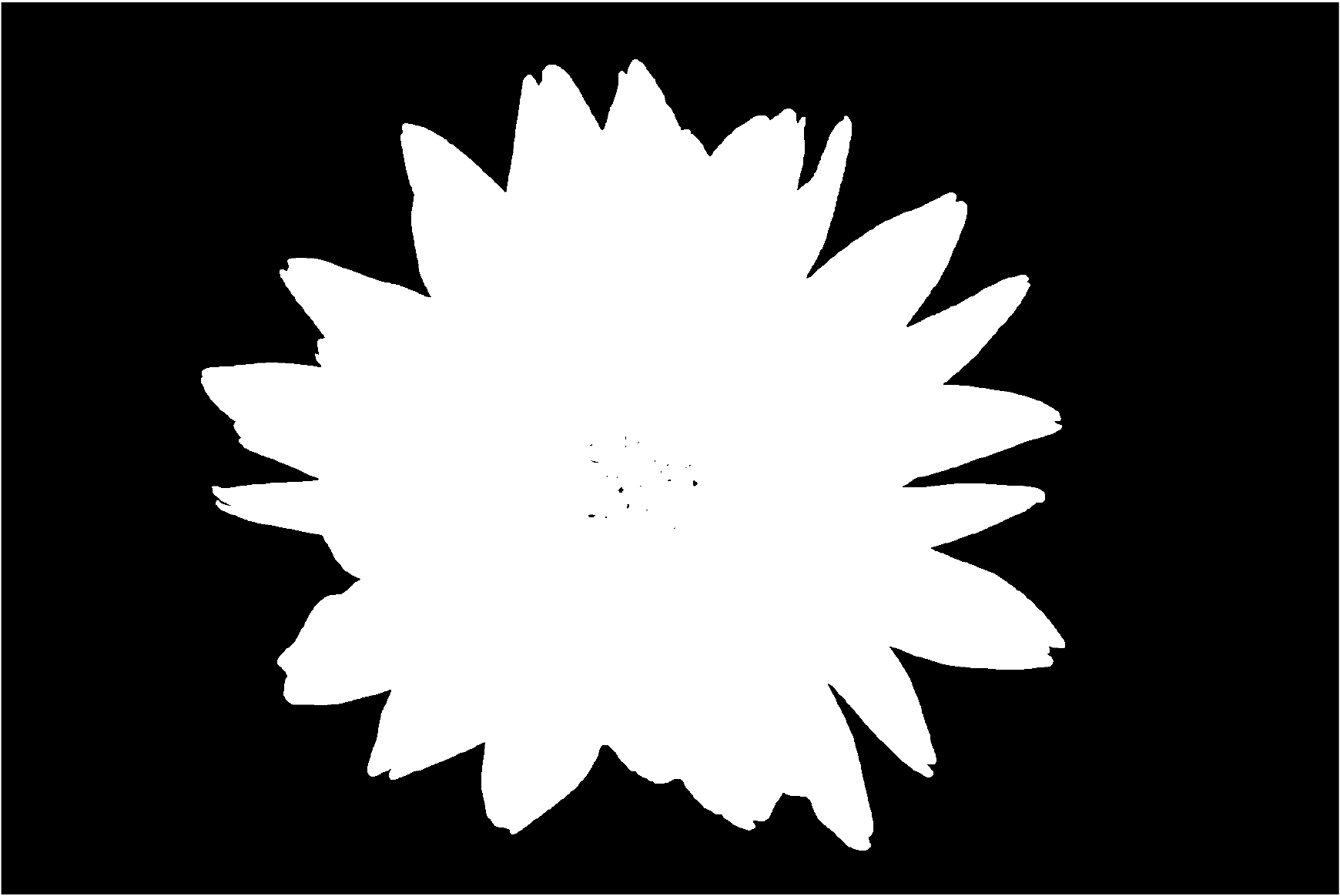}\\
             b) Graph cuts}
        \hspace{-0.25em}
        \shortstack{
             \includegraphics[width=0.175\linewidth]{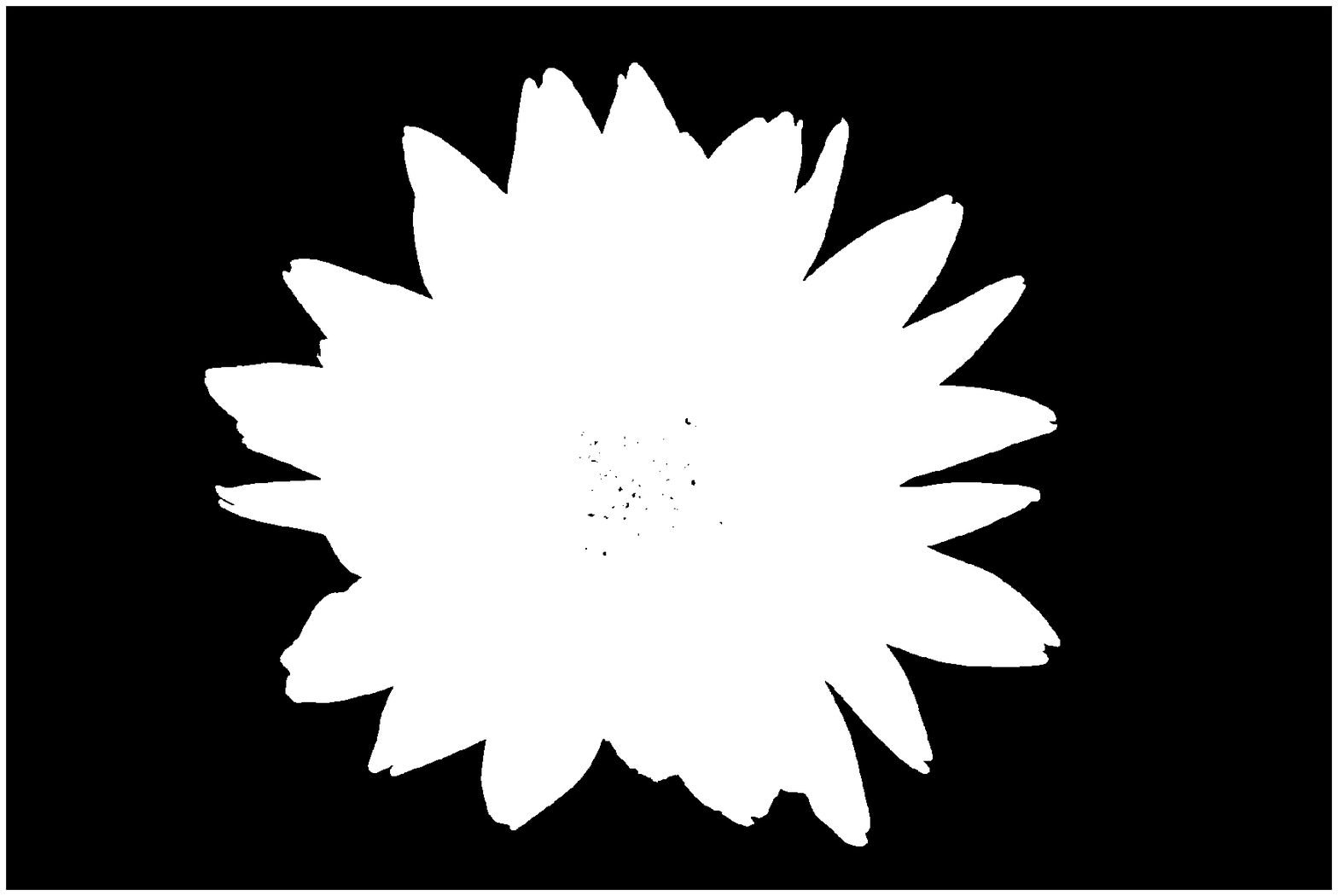}\\
             c) Our \dope{} method}
            \includegraphics[width=0.44\linewidth]{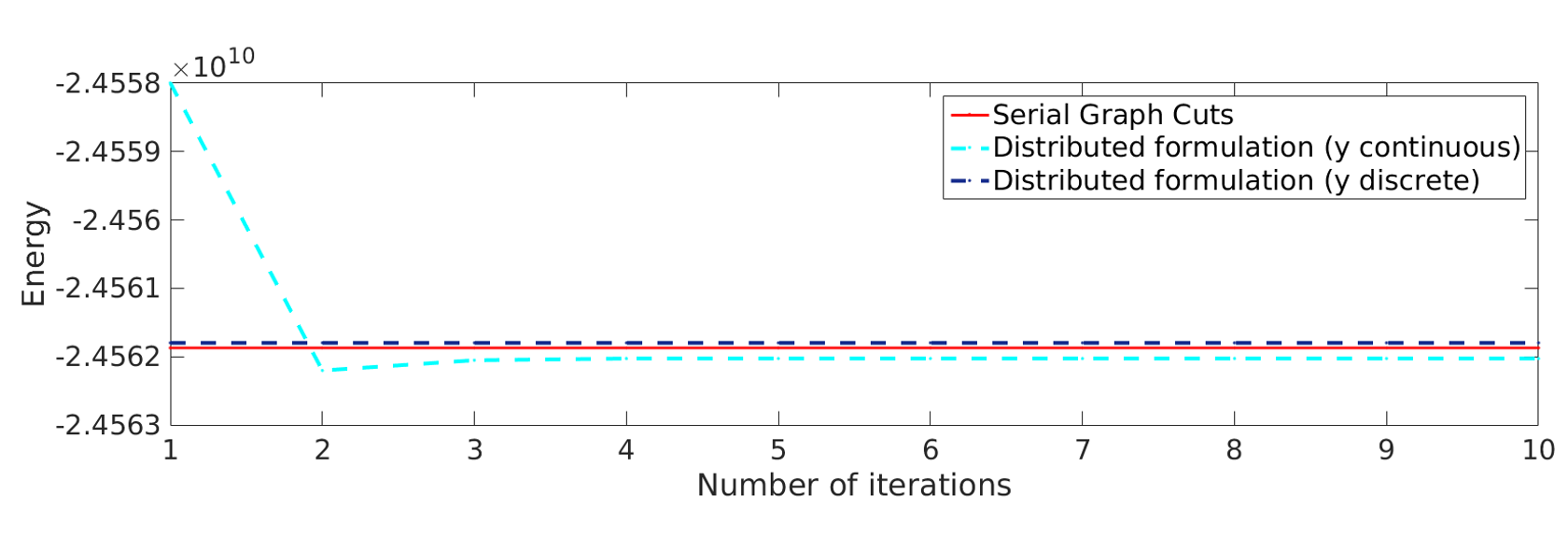}
        }
        
         \mbox{
        \shortstack{
             \includegraphics[width=0.175\linewidth]{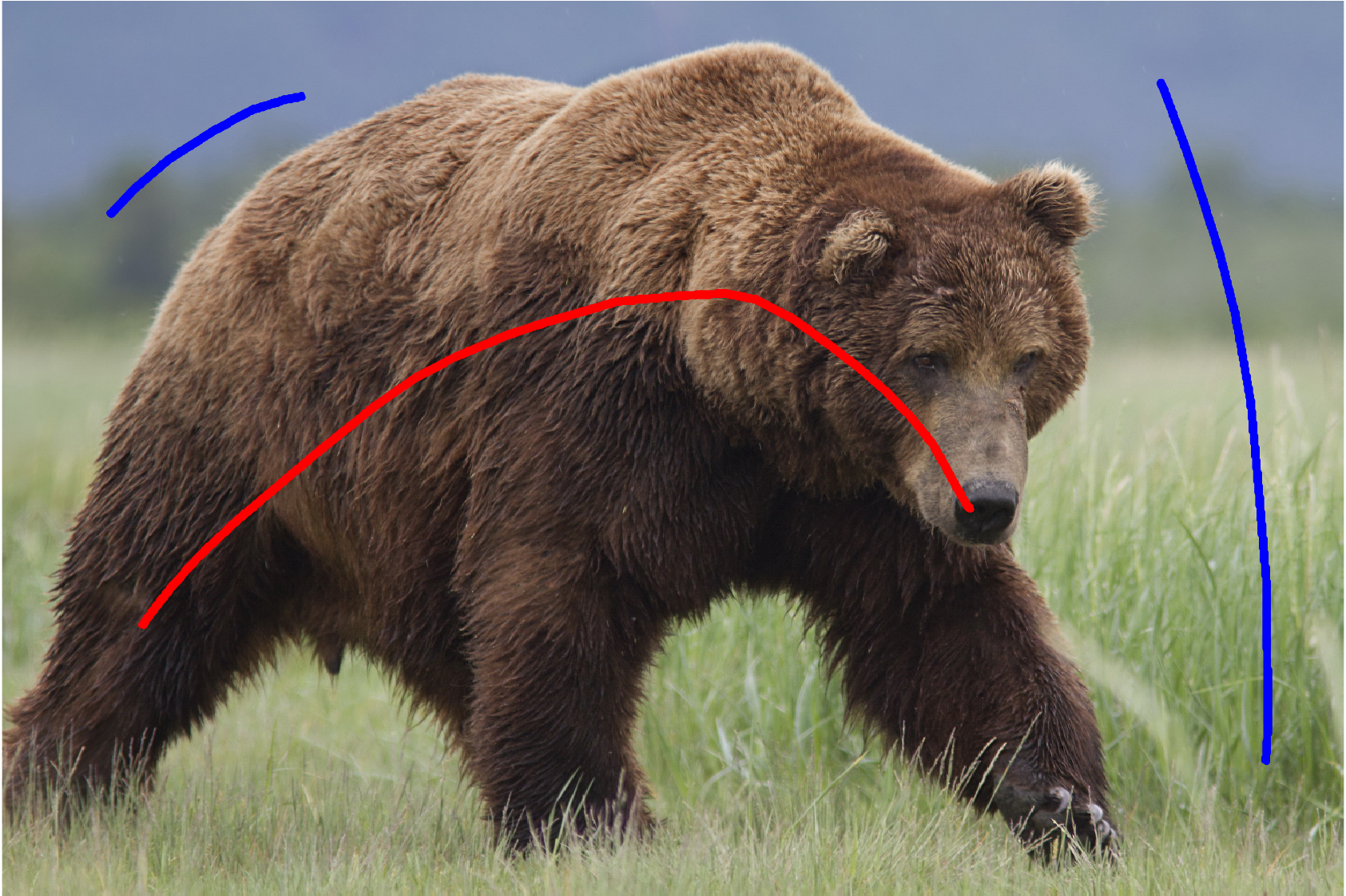}\\
            a) Original image }
        \hspace{-0.25em}
        \shortstack{
             \includegraphics[width=0.175\linewidth]{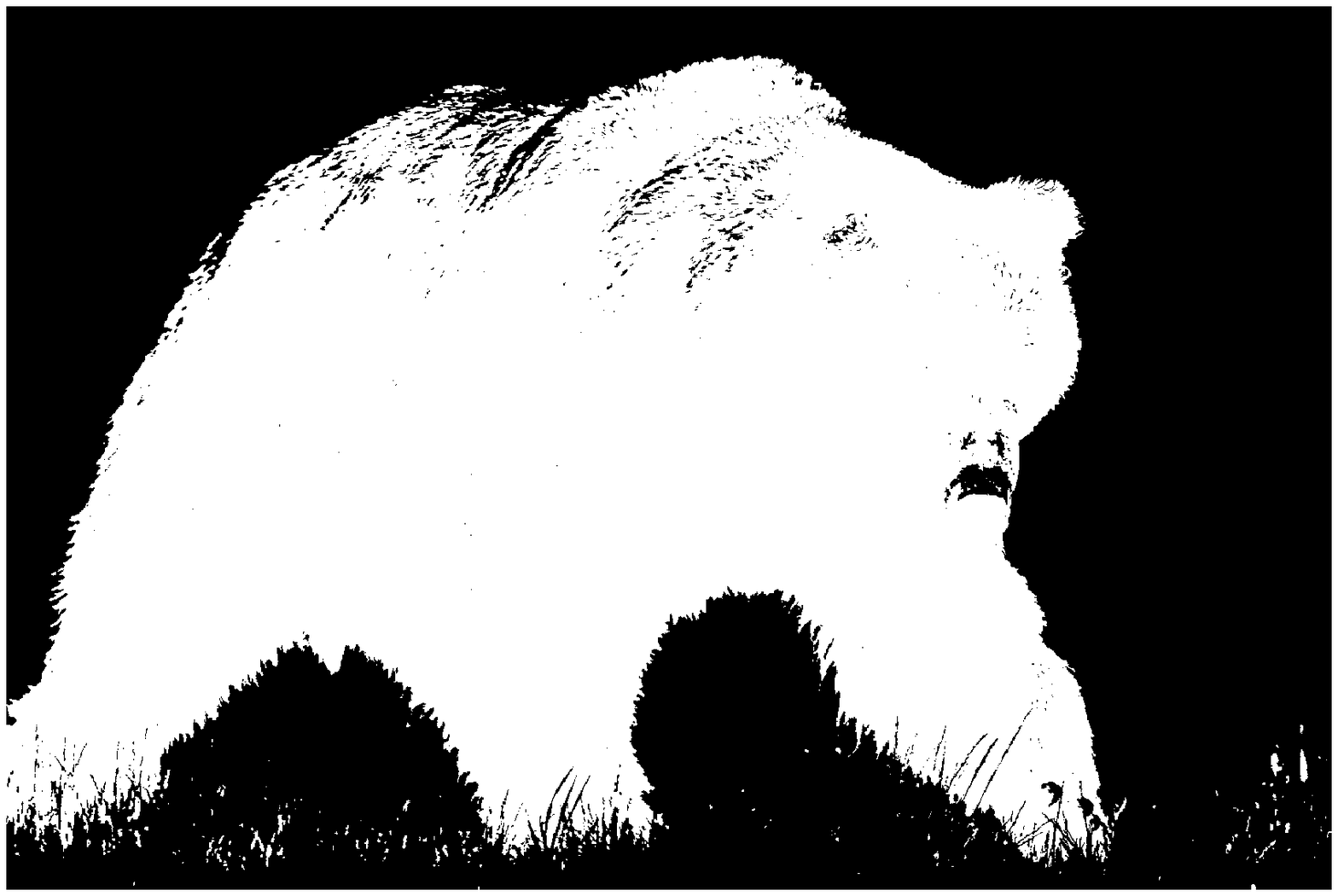}\\
             b) Graph cuts}
        \hspace{-0.25em}
        \shortstack{
             \includegraphics[width=0.175\linewidth]{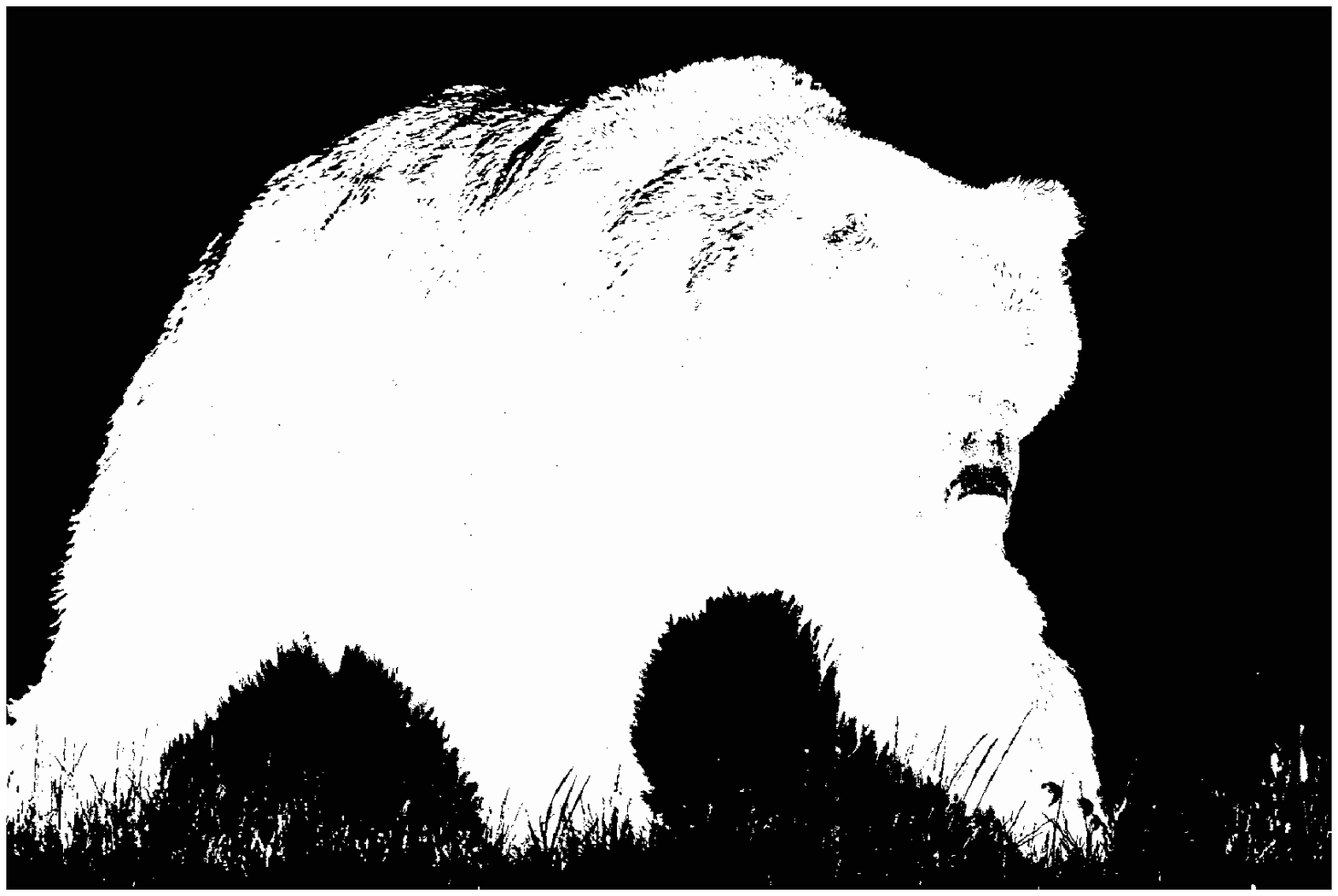}\\
             c) Our \dope{} method}
             \shortstack{
            \includegraphics[width=0.44\linewidth]{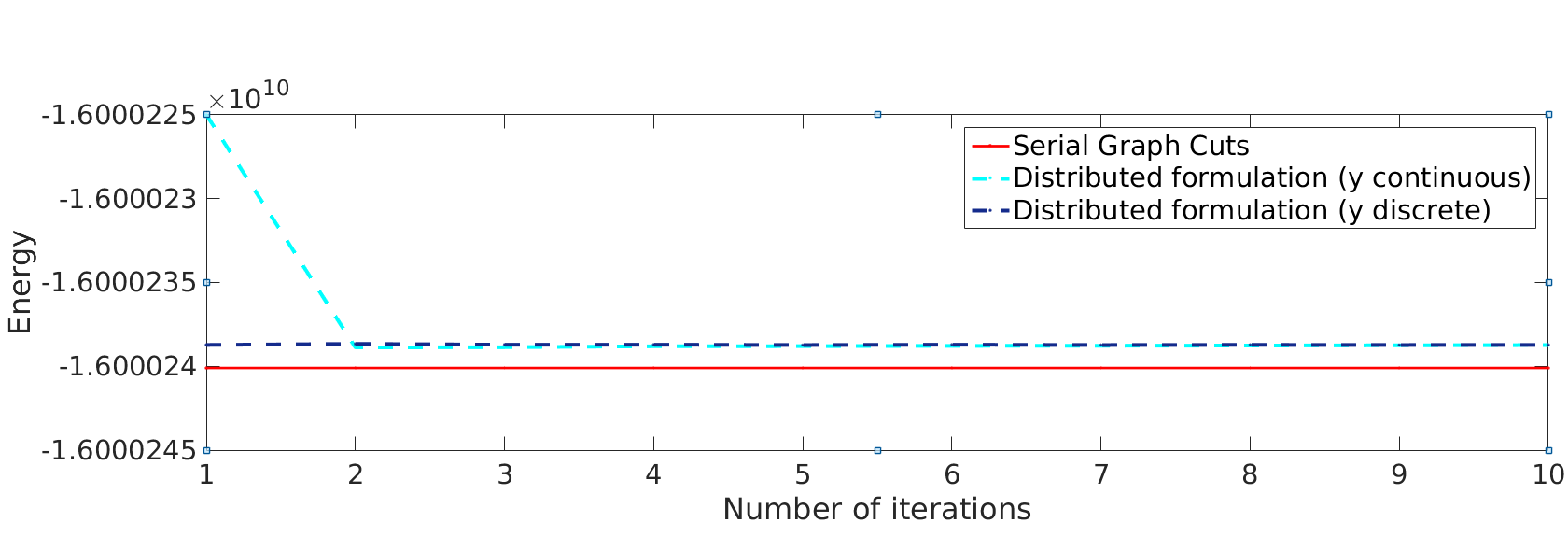}\\
            d) Energy evolution}
        }
    \end{center}
    \caption{Visual example of 2D image segmentation by sGC and \dope{} approaches (\emph{left}) and the evolution of the segmentation energy in both formulations with respect to the number of iterations (\emph{right}). }%
   \label{fig:VisualExamples}
\end{figure*}

We first tested our method on 10 high-resolution 2D multi-channel images, with resolution ranging between 2000$\times$3000 and 2600$\times$3900 pixels. As in \cite{Boykov2006}, we drew seeds to generate color model priors for the foreground and background regions (see Fig. \ref{fig:VisualExamples}). The k-means algorithm \cite{macqueen1967some} was employed to group foreground/background seed pixels into $5$ clusters, which were then used to compute the log posteriors of unseeded pixels.

\begin{figure}[htb!]
     \begin{center}
        \vspace{-.5\baselineskip}
        
        \mbox{%
           \shortstack{\scriptsize{Num blocks = 128} \\ \includegraphics[width=0.23\textwidth]{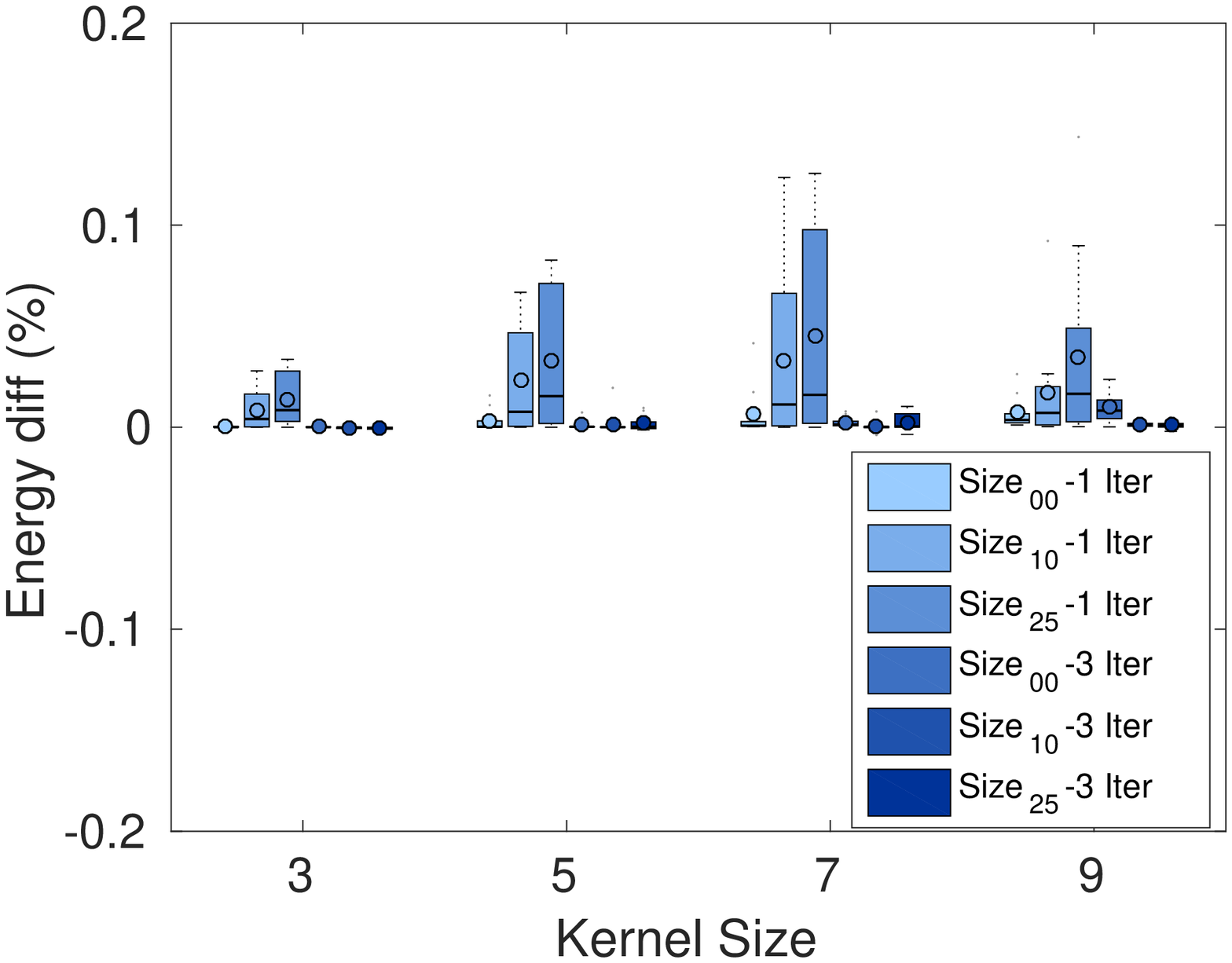}}

            \hspace{0.5em}
        
        \shortstack{\scriptsize{$\mr{Size}_{25}$} \\ \includegraphics[width=0.23\textwidth]{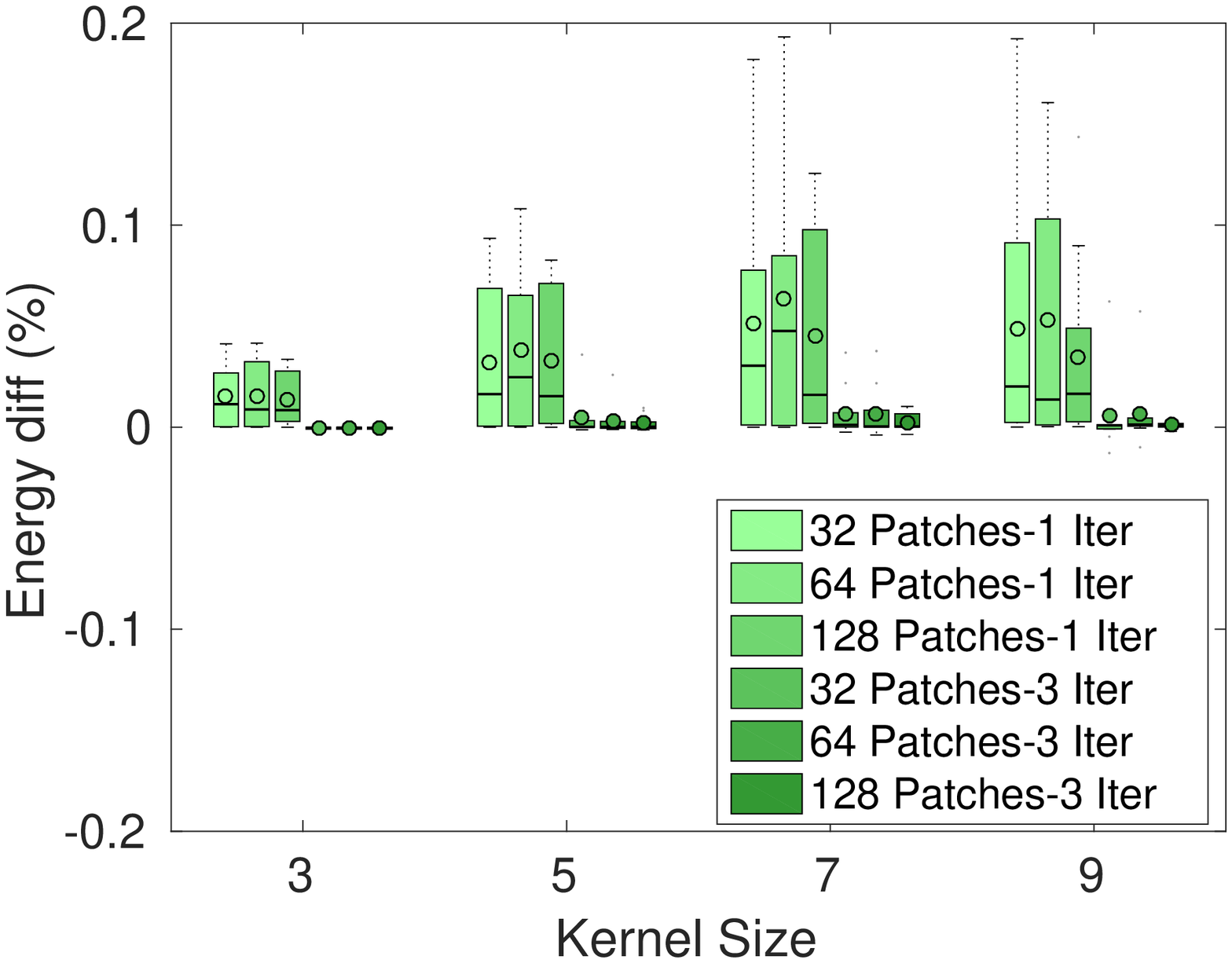}}
        }%
        
        \vspace{1mm}
        
        \mbox{
            \includegraphics[width=0.23\textwidth]{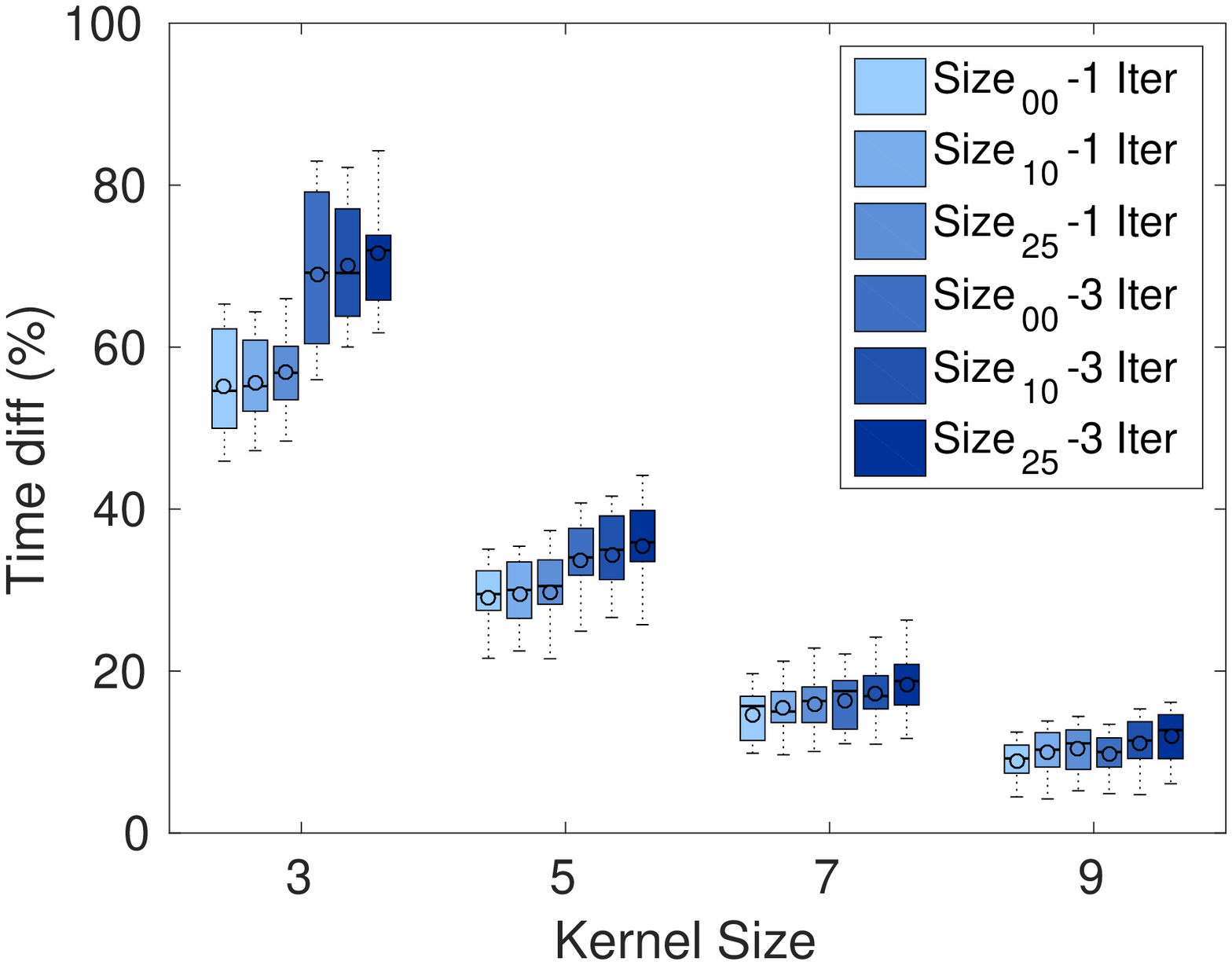}
             
            \hspace{0.5em}
        
            \includegraphics[width=0.23\textwidth]{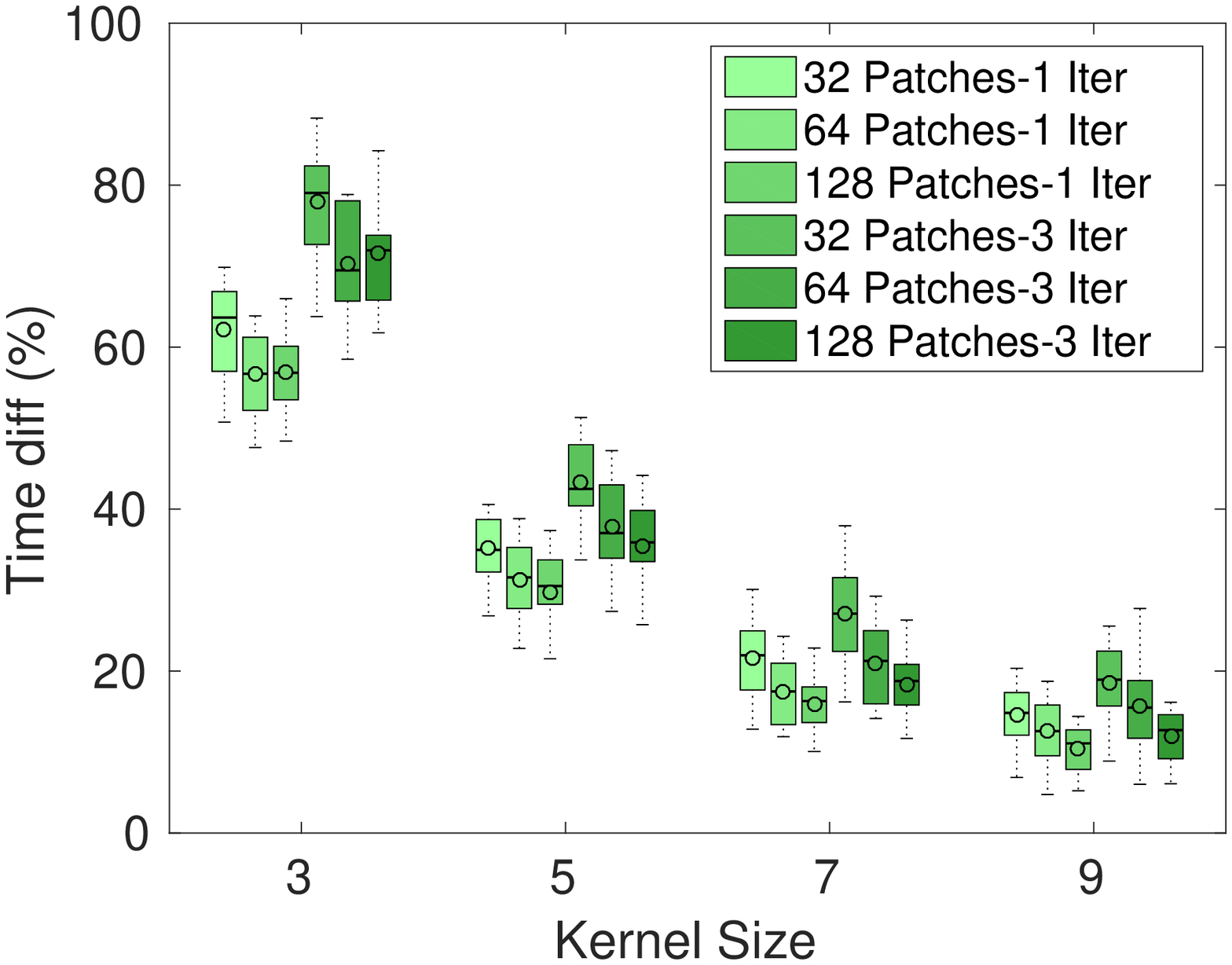}
        }
        
        \vspace{1mm}
        
        \mbox{%
            
            \includegraphics[width=0.23\textwidth]{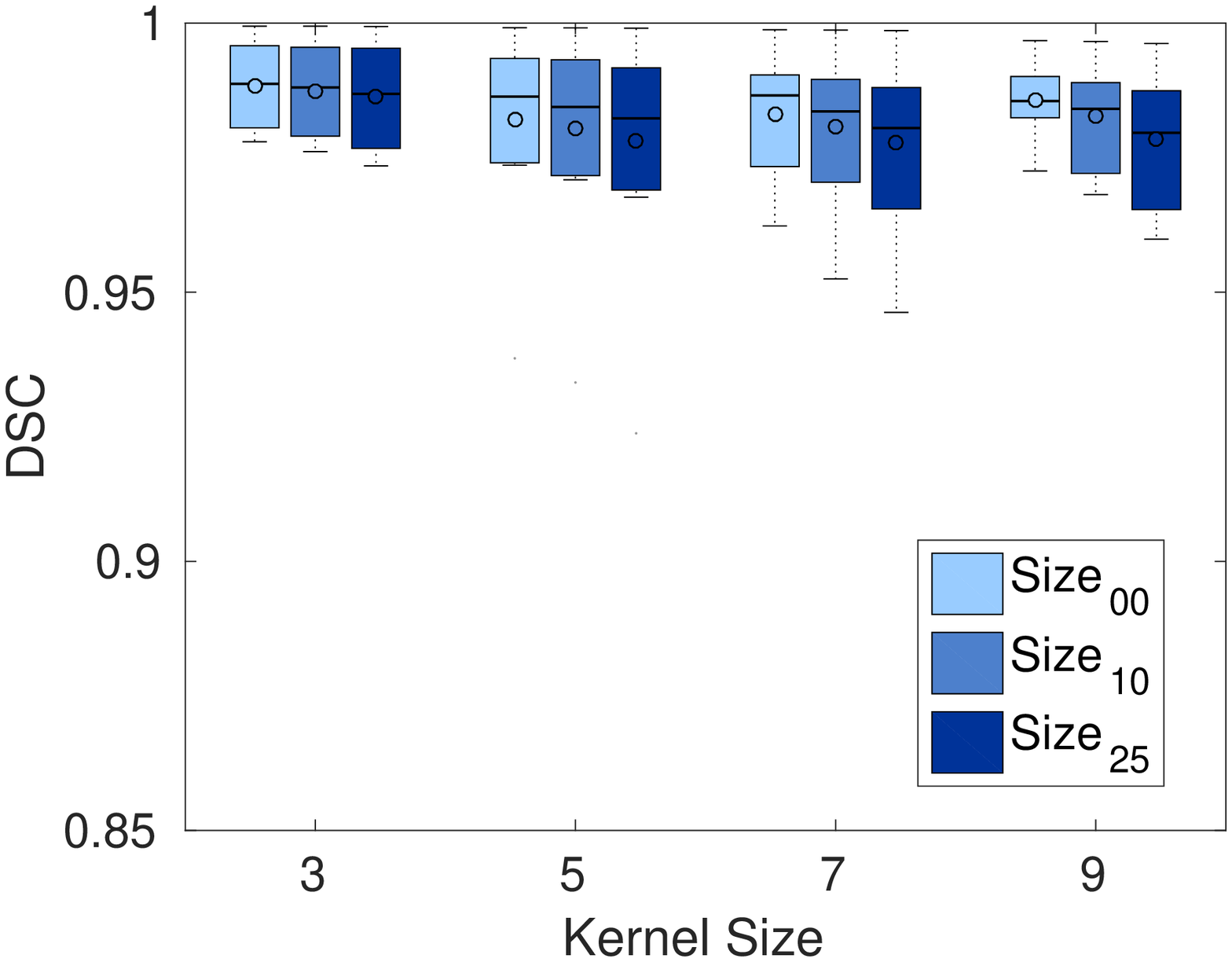}
        
         \hspace{0.5em}
            
            \includegraphics[width=0.23\textwidth]{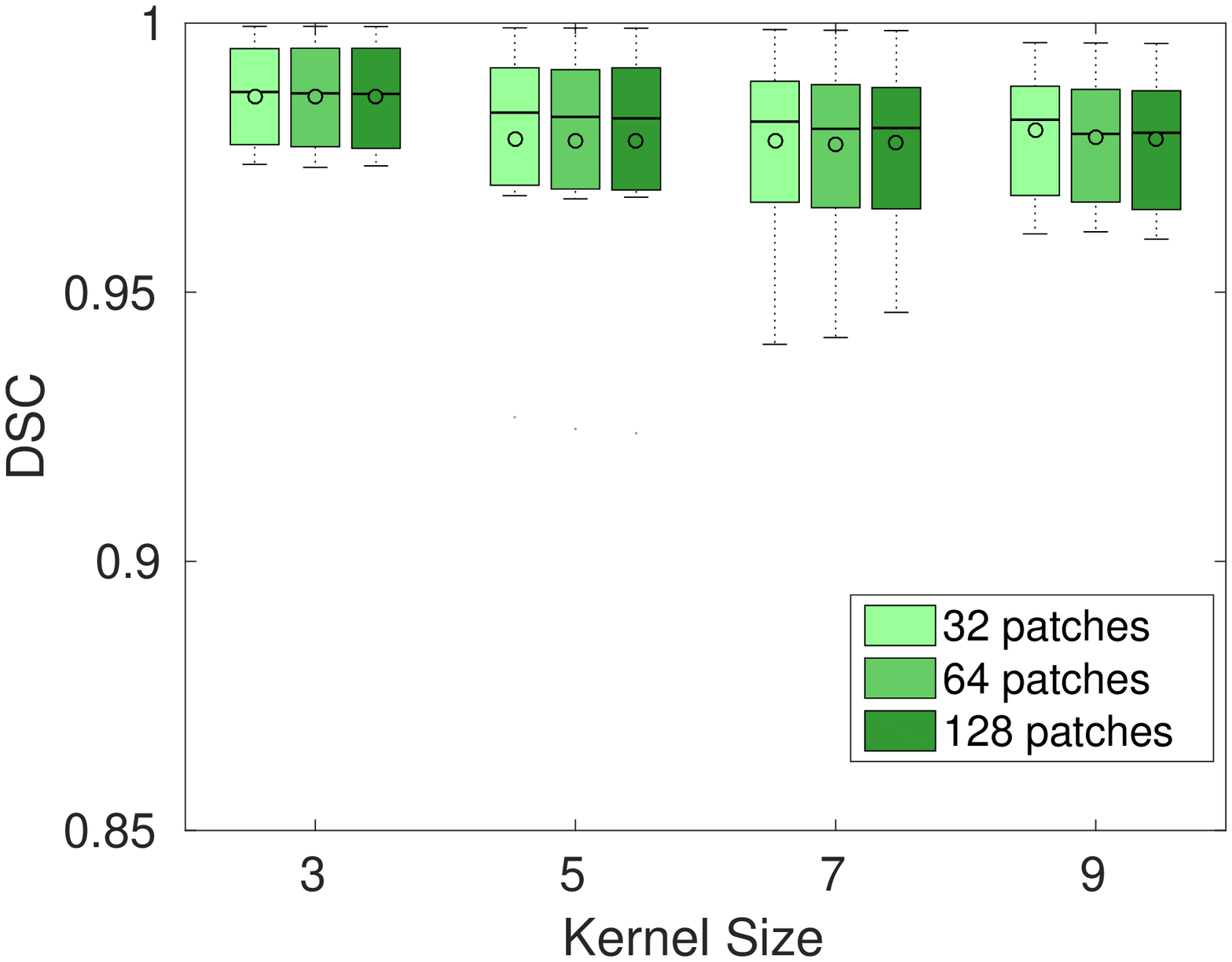}
        }
             
    \end{center}
    \caption{Mean relative energy difference (\%), relative time difference (\%) and Dice similarity coefficient (DSC) between the segmentation of sGC and our \dope{} method, computed over the 10 high-definition 2D images. In the left-side plots, the number of blocks is fixed to 128, and values are reported for different block and kernel sizes. In right-side plots, the block size is fixed to $\mr{Size}_{25}$, and values are reported for different number of blocks and kernel sizes.}%
   \label{fig:energyTimeDSC}
\end{figure}

Figure \ref{fig:energyTimeDSC} gives the average relative energy difference (\%), relative time difference and Dice similarity coefficient (DSC) between the segmentation of our \dope{} method and sGC, computed over the 10 images. In the left-side plots, the number of blocks was set to 128, but we varied the size of these blocks and the kernel. Conversely, the right-side plots compare our method with sGC for various numbers of blocks and kernel sizes, while keeping the block size fixed to $\mr{Size}_{25}$. Values are reported for one and three ADMM iterations of our method. While a more detailed analysis is presented below, we found that three iterations were often sufficient to achieve convergence in the case of 2D images (e.g., see Fig. \ref{fig:VisualExamples}). 
We observe that energy differences are quite small in all tested configurations, with values around 0.1\% for one ADMM iteration and 0.01\% for three ADMM iterations. With respect to block size, we observe no difference between the tested configurations, for the same number of ADMM iterations. For a single ADMM iteration, increasing the overlap seems to result in higher energy differences. However, these differences disappear when using three iterations, suggesting that having a greater overlap requires more iterations to converge.

As expected, segmentation times varied proportionally to the number and size of blocks. However, doubling the number of blocks did not lead to reduction in processing time by the same factor. This is in part due to pre-processing operations, such as computing the unary and pairwise potentials for the whole image, which need to be performed regardless of the image partitioning scheme used. Another trend that can be observed is that the speed-up provided by our method increases with the kernel size. Thus, for kernel sizes of 7 or more, our method obtained nearly identical segmentation results up to 5 times faster than sGC. Additionally, allowing the algorithm to run three iterations did not increase run times significantly for larger kernels, suggesting that most operations are performed in pre-processing steps. 

In terms of segmentation consistency, it can be seen that our \dope{} method obtains segmentation results quite similar to those of sGC, with DSC values above 0.99. In most cases, increasing the number of blocks decreases DSC values, although this difference is not significant. A similar effect can be observed when employing larger blocks and kernels. Overall, the segmentation results obtained by our method are consistent with those of sGC, for all tested configurations. 


Figure \ref{fig:VisualExamples} gives two examples of segmentations obtained by sGC and our \dope{} method. The first column shows the image to be segmented with foreground/background scribbles, whereas the second and third columns give the segmentation result of sGC and our method, respectively. The evolution of the segmentation energy is also shown in Figure \ref{fig:VisualExamples} (\emph{right}). We observe that our approach converges rapidly, requiring only two iterations to achieve near-zero energy differences. 


\subsection{3D MRI volumes}
\label{ssection:3D}

\begin{figure}[htb!]
     \begin{center}
     \footnotesize{\textbf{ADMM energy}}
     \mbox{
        \shortstack{
        \includegraphics[width=1\linewidth]{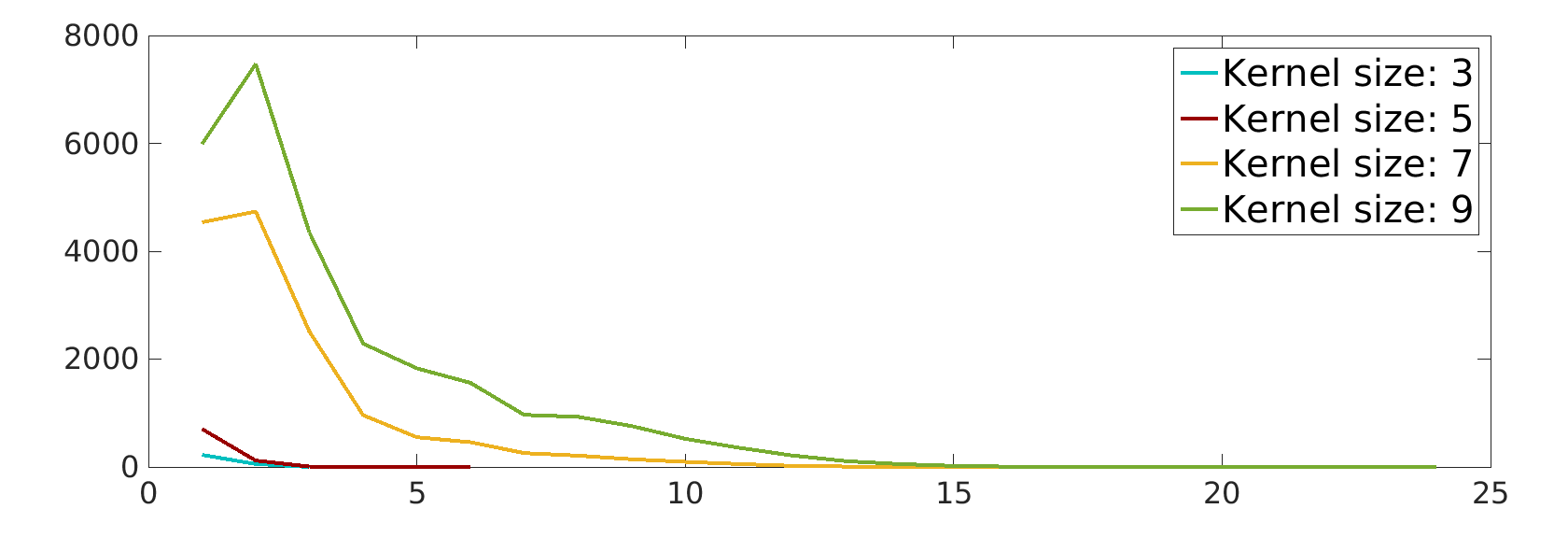} \\[-1mm]
        \footnotesize{Number of iterations}
        }
    } 
            
    \caption{Evolution of the ADMM energy (augmented Lagrangian terms) for a partitioning comprised of 128 blocks with $\mr{Size}_{10}$ and different kernel sizes.}
    
   \label{fig:evolutionADMM}
   \end{center}  
\end{figure}


\begin{figure}[htb!]
     \begin{center}
     \mbox{
        \includegraphics[width=0.99\linewidth]{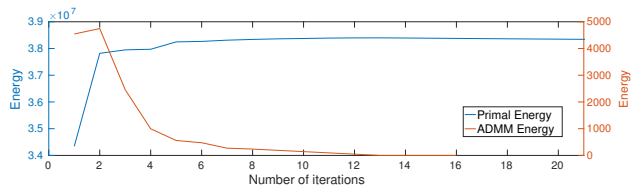}}
    \caption{Evolution of the segmentation (primal) energy and ADMM (augmented Lagrangian terms) energy during optimization, for a partitioning comprised of 128 blocks with $\mr{Size}_{10}$ and a kernel of size 7.}
   \label{fig:evolution}
   \end{center}  
\end{figure}

Segmentation efficiency is particularly important in the case of 3D volumes, where computational and memory requirements often exceed the capacity of current methods. As a second experiment, we tested our \dope{} method on a 3D MRI brain volume of size 200$\times$200$\times$100. For this experiment, we considered the task of segmenting sub-cortical brain regions, and used the soft probability map generated by a 3D convolutional neural network \cite{dolz20163dFCNN} \footnote{\url{https://github.com/josedolz/LiviaNET}} as unary potentials in the energy function.

\begin{figure*}[ht!]
     \begin{center}
     \shortstack{
        \includegraphics[width=0.16\linewidth]{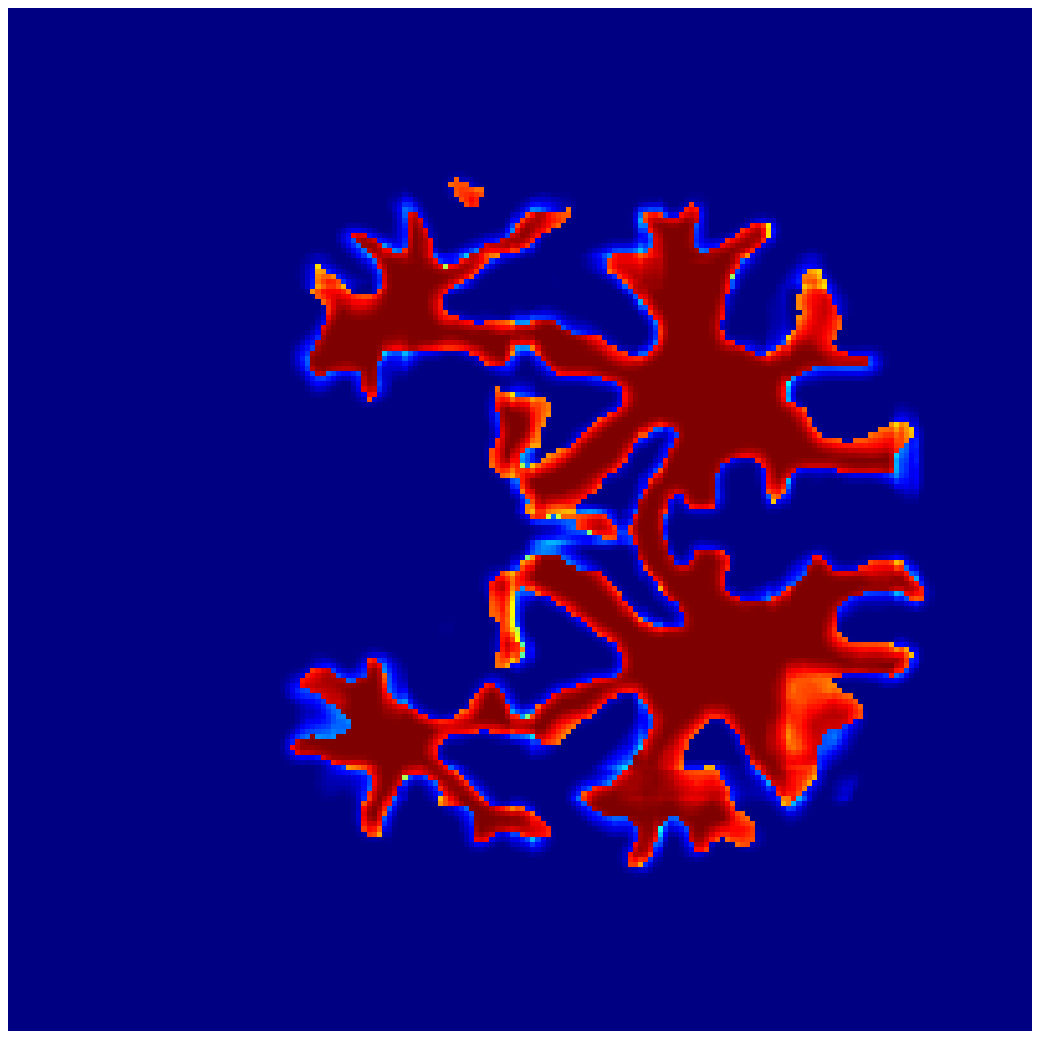}\\
        1 iteration}
    \shortstack{
        \includegraphics[width=0.16\linewidth]{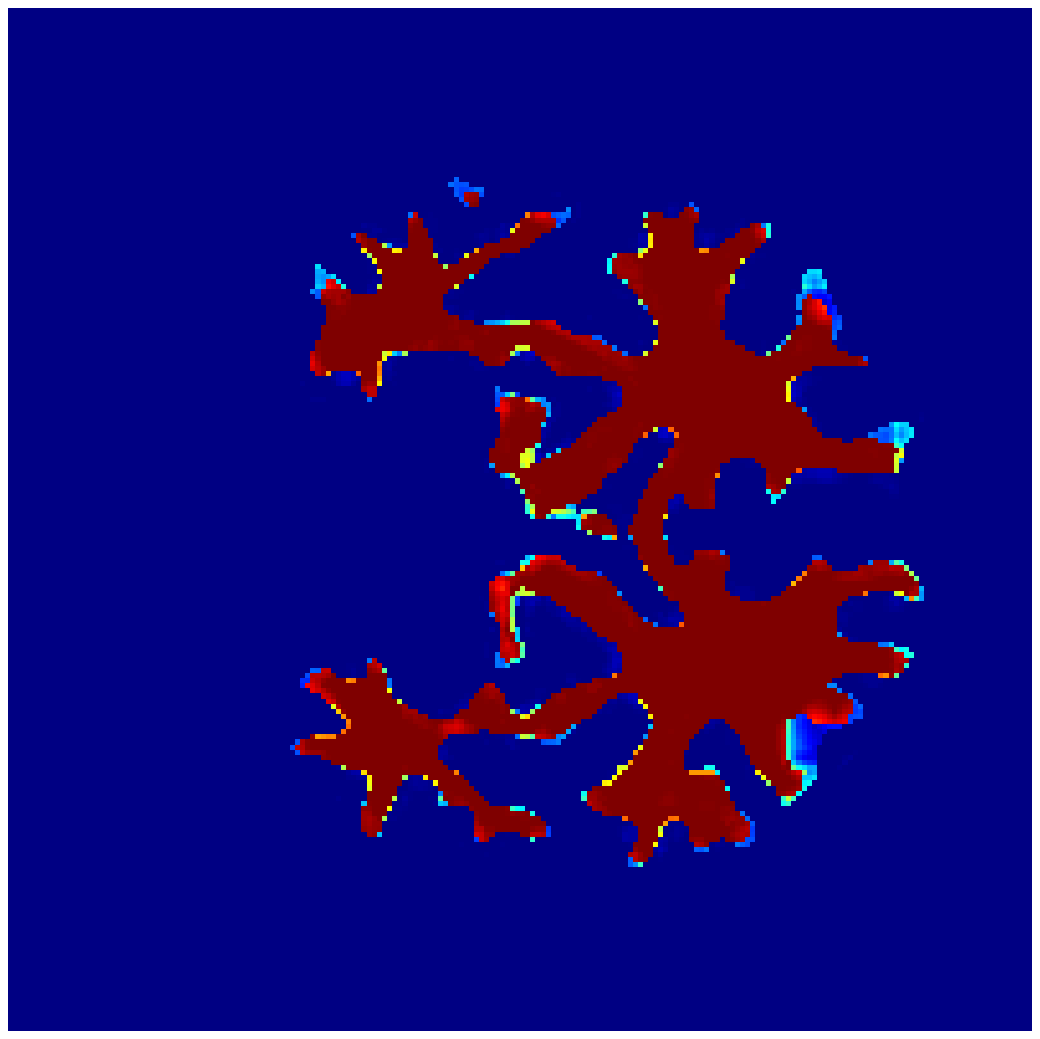}\\
        2 iterations}
    \shortstack{
        \includegraphics[width=0.16\linewidth]{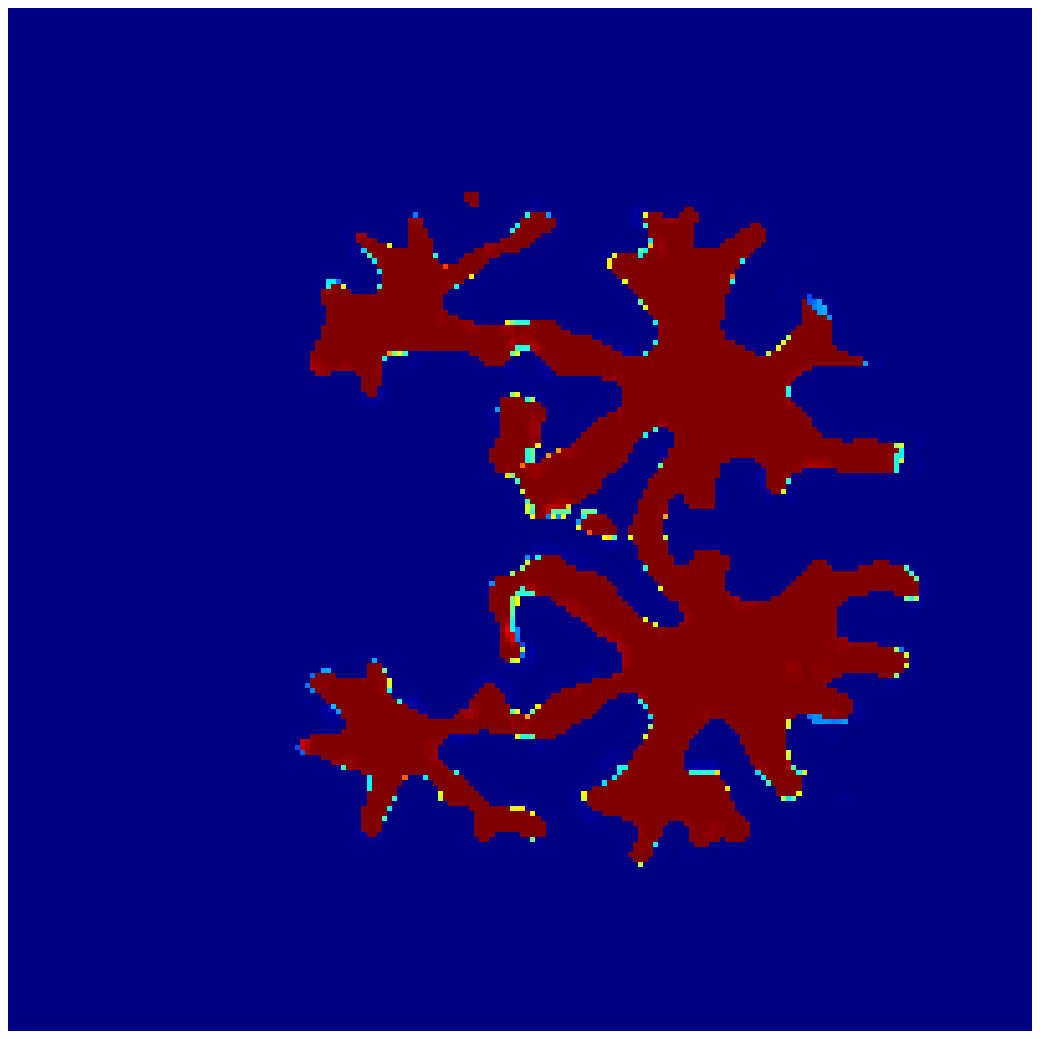}\\
        3 iterations}
    \shortstack{
        \includegraphics[width=0.16\linewidth]{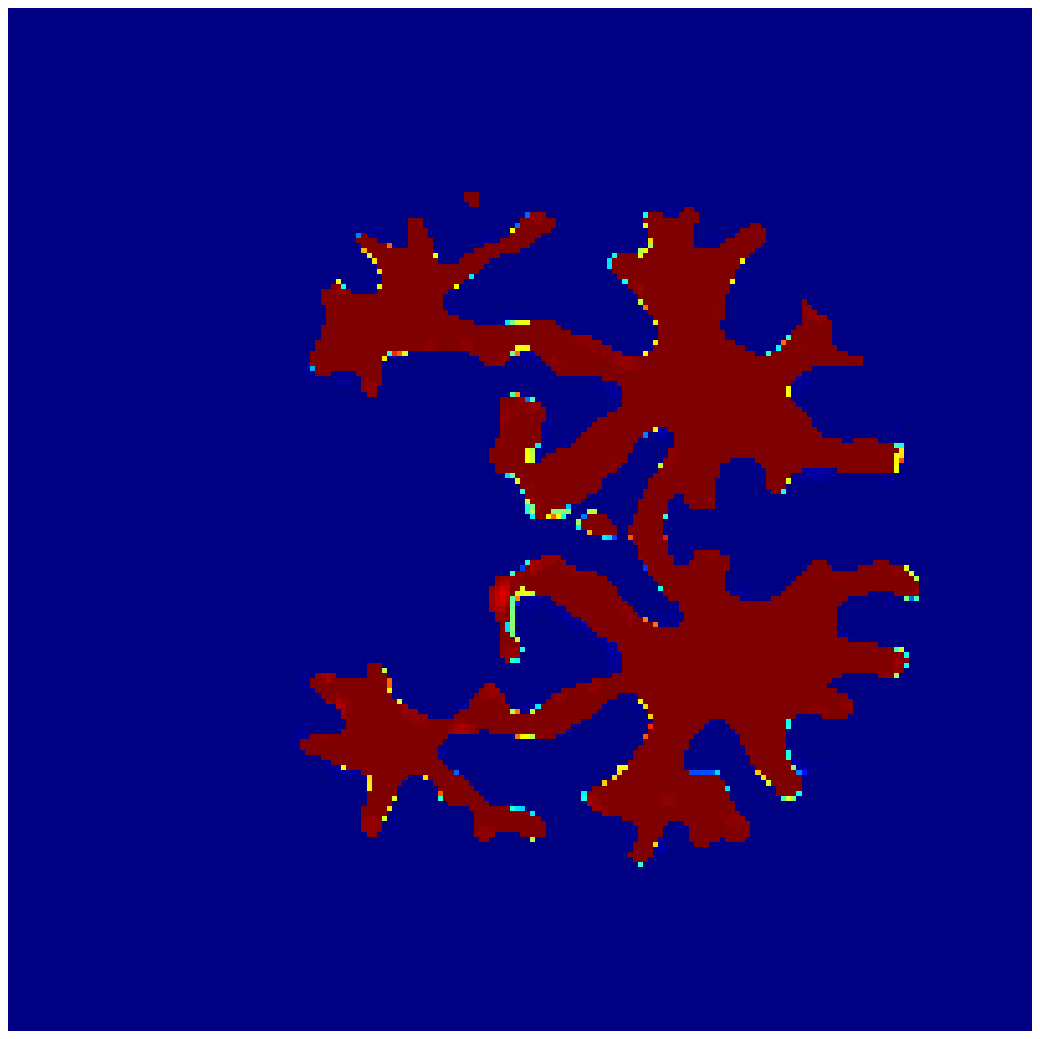}\\
        4 iterations}
    \shortstack{
        \includegraphics[width=0.16\linewidth]{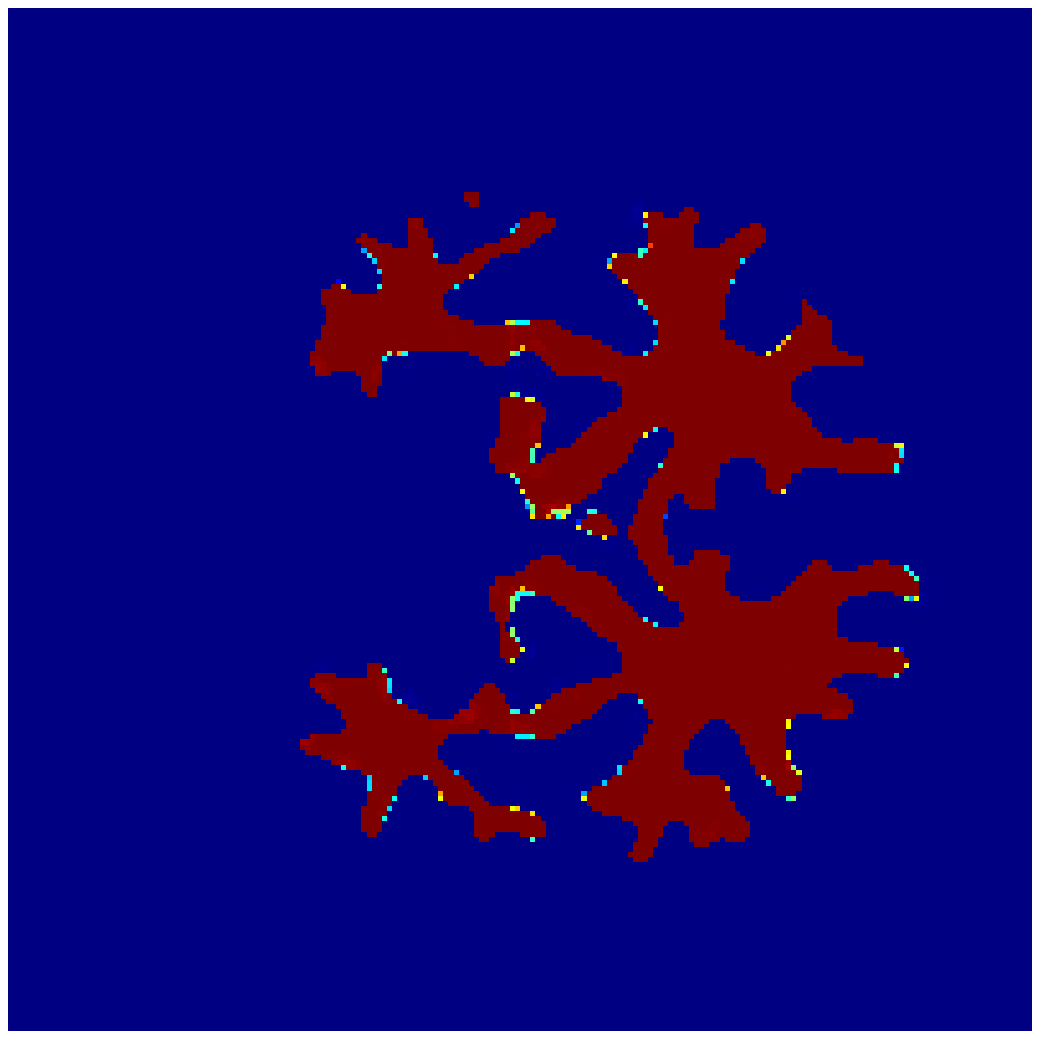}\\
        5 iterations}
    \shortstack{
        \includegraphics[width=0.16\linewidth]{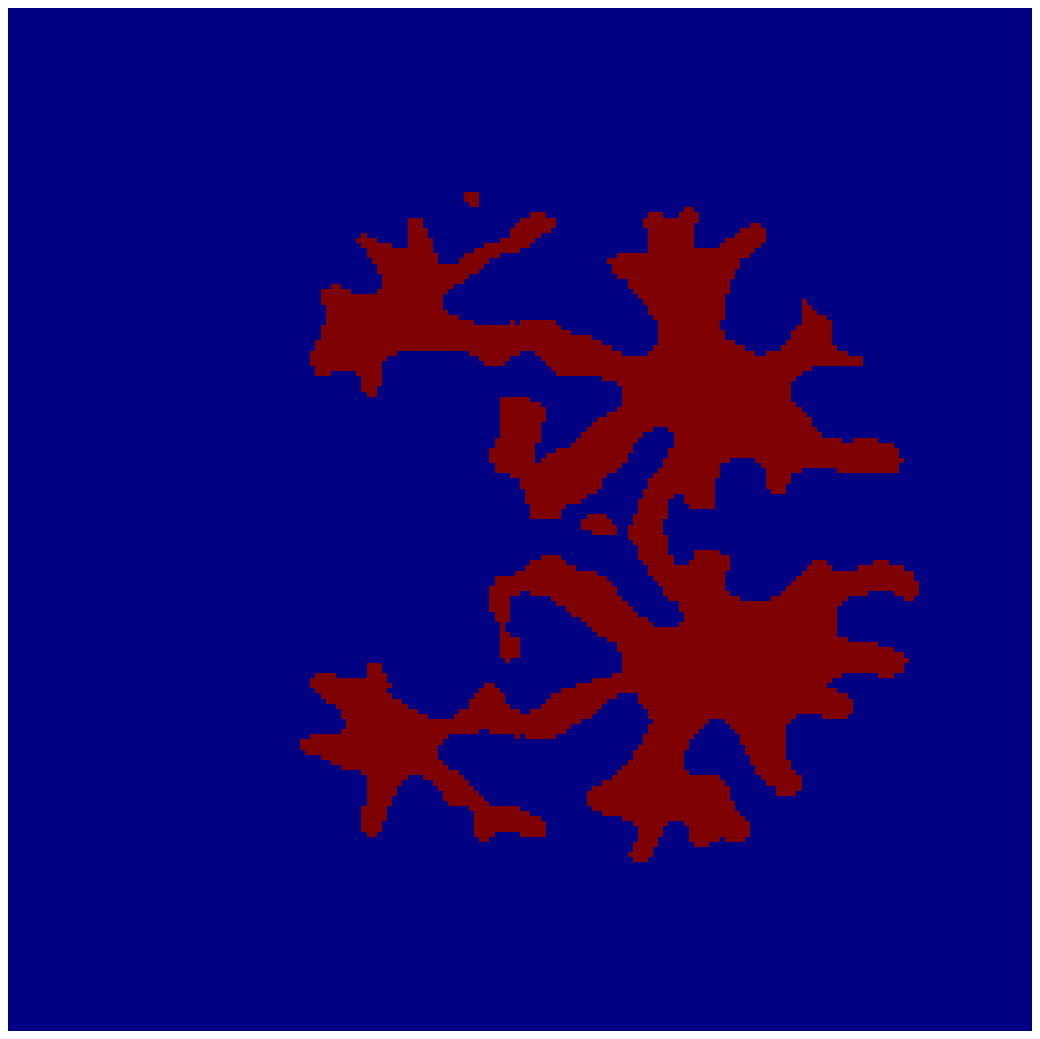}\\
        22 iterations}
    \caption{Evolution of the segmentation results (relaxed $\yy$) with respect to the number of iterations.}
    
   \label{fig:SegevolutionJet}
   \end{center}  
\end{figure*}

Figure \ref{fig:evolutionADMM} shows the energy related to the augmented Lagrangian terms in Eq. (\ref{eq:admm-formulation}), for a partitioning composed of 128 blocks with $\mr{Size}_{10}$, and kernel sizes corresponding to 3, 5, 7 and 9. Recall that this energy corresponds to segmentation consistency across different blocks.
We observe that the number of iterations required to achieve convergence increases with kernel size, probably due to the broader interaction between blocks for larger kernels. However, our method converged in less than 10 iterations, for all kernel sizes. These observations are confirmed by Figure \ref{fig:evolution}, which also shows the variation of segmentation energy (unary and pairwise potentials) when employing 128 blocks with $\mr{Size}_{10}$ and a kernel of size 7. We notice that the segmentation energy increases with the number of iterations. This can be explained by the fact that this energy is computed using the integer-relaxed segmentation vector $\yy$, which gets increasingly restricted to a binary solution over time. Segmentation convergence is illustrated in Fig. \ref{fig:SegevolutionJet}, which shows the evolution of $\yy$ for a random 2D slice of the volume.  

Analyzing detailed results, we see that mean relative energy differences increase with kernel size, ranging from 0.1$\%$ for kernels of size 3 to 2.5$\%$ when employing kernel of size 9. Moreover, for a fixed kernel size, having a greater overlap leads to smaller energy differences (e.g., energy difference of 1.5$\%$ for $\mr{Size}_{20}$ compared to 2.25$\%$ for $\mr{Size}_{00}$, for a kernel size of 9 and 128 blocks). This suggests the greater usefulness of having overlapping blocks in the segmentation of 3D volumes. However, when overlap is allowed, larger block sizes reported slightly higher energy differences. As was the case for 2D image segmentation, the speed-up of our \dope{} method depends on kernel size and the block number/size. Hence, for 64 blocks with $\mr{Size}_{10}$ and kernel size of 7, our method was about 3 times faster than sGC, while a speed-up of $4$ was achieved using 128 blocks with $\mr{Size}_{00}$ and a kernel size of 9. For segmentation consistency, DSC values ranged from 0.95 to 0.99 in all tested configurations. While a few iterations were sufficient for small kernels, larger kernels required more iterations to converge. 



\begin{figure}[ht!]
     \begin{center}
     \mbox{
        \includegraphics[width=0.32\linewidth,height=0.32\linewidth]{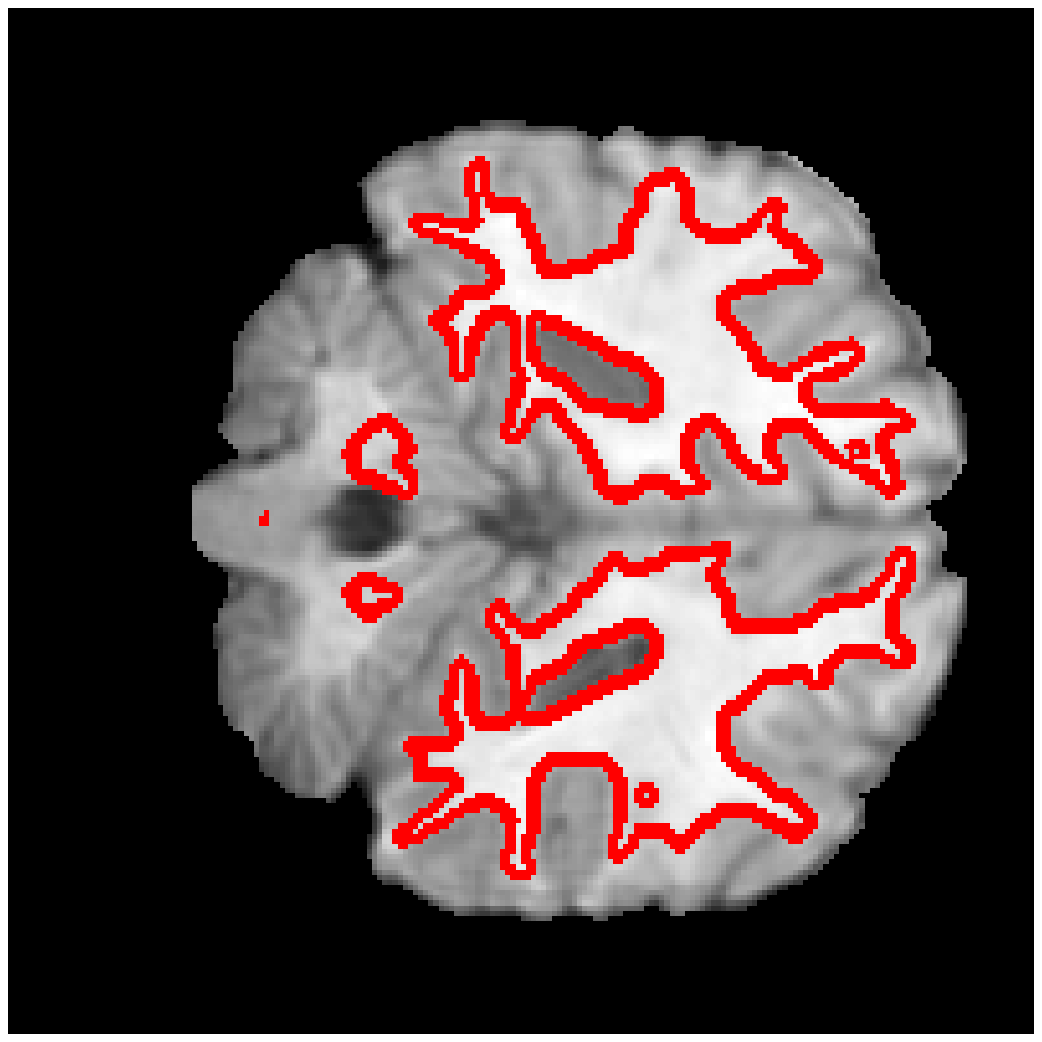}
        
         \hspace{-.5 mm}
         
        \includegraphics[width=0.32\linewidth,height=0.32\linewidth]{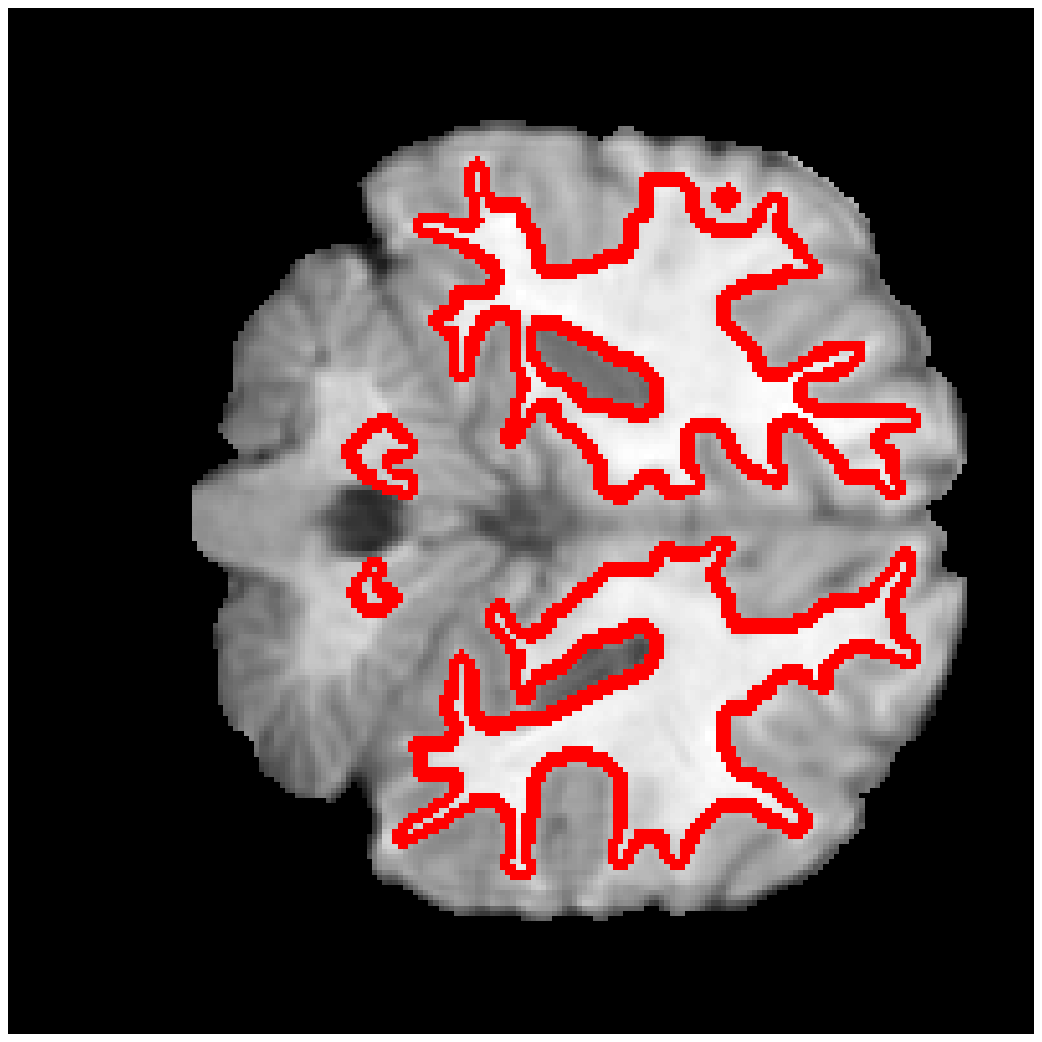}
        
         \hspace{-.5 mm}
         
        \includegraphics[width=0.32\linewidth,height=0.32\linewidth]{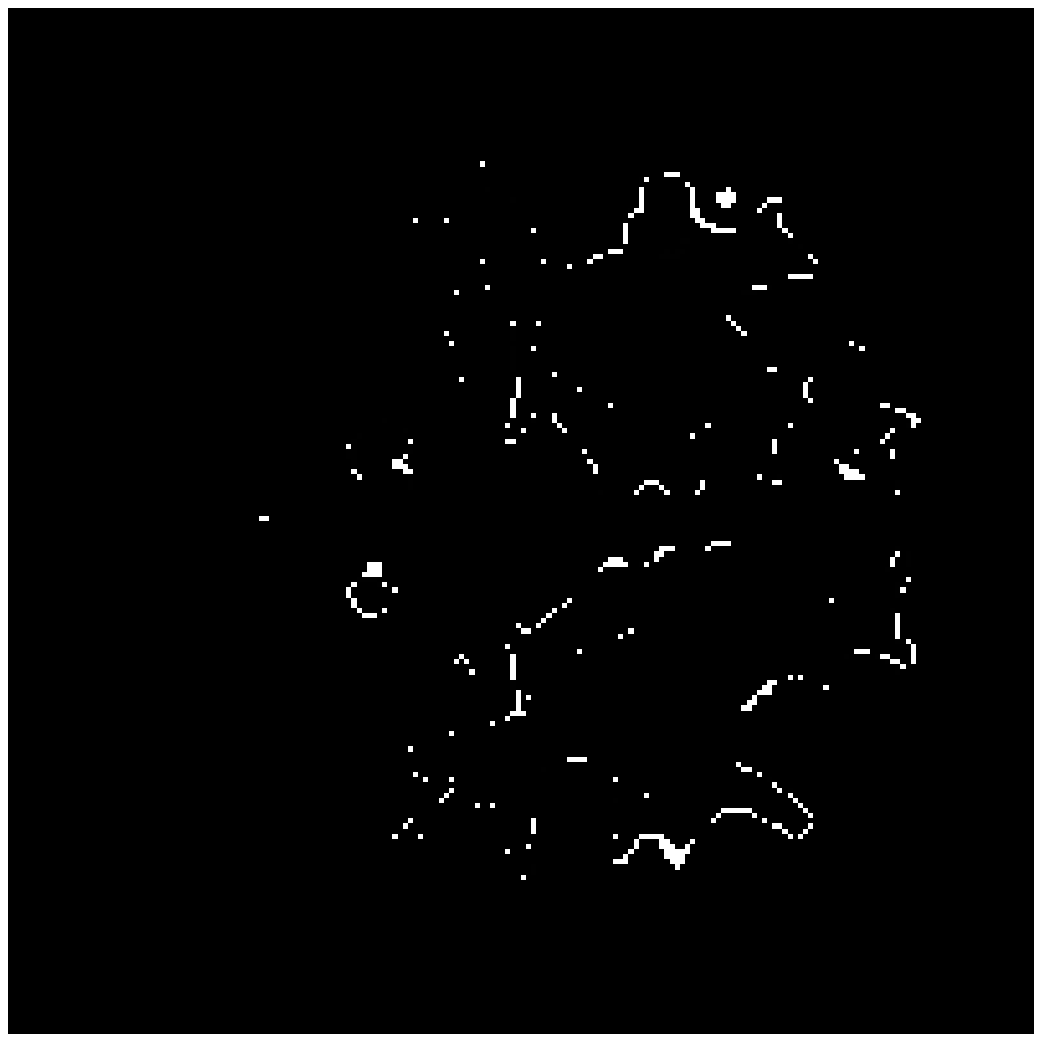}
        }
     \mbox{
        \includegraphics[width=0.32\linewidth,height=0.32\linewidth]{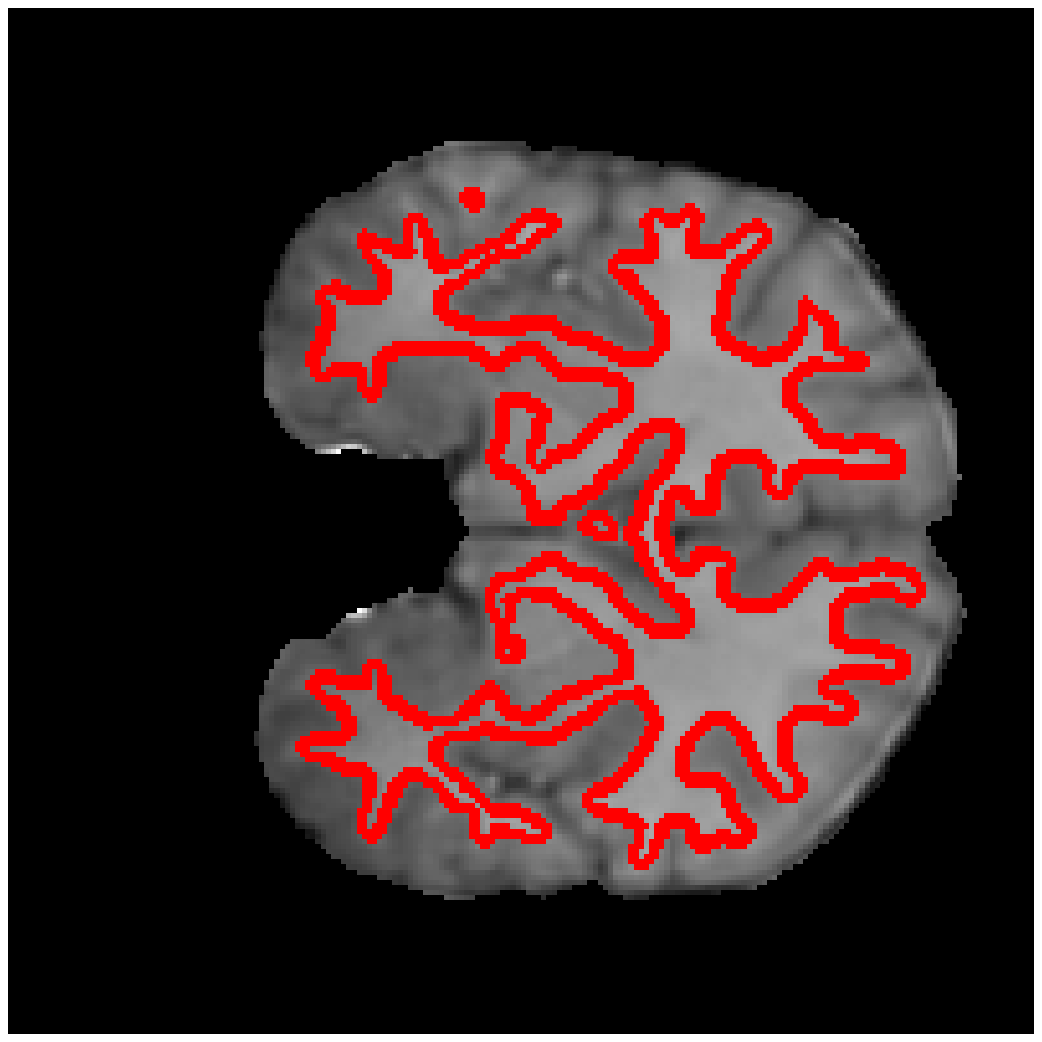}
        
         \hspace{-.5 mm}
         
        \includegraphics[width=0.32\linewidth,height=0.32\linewidth]{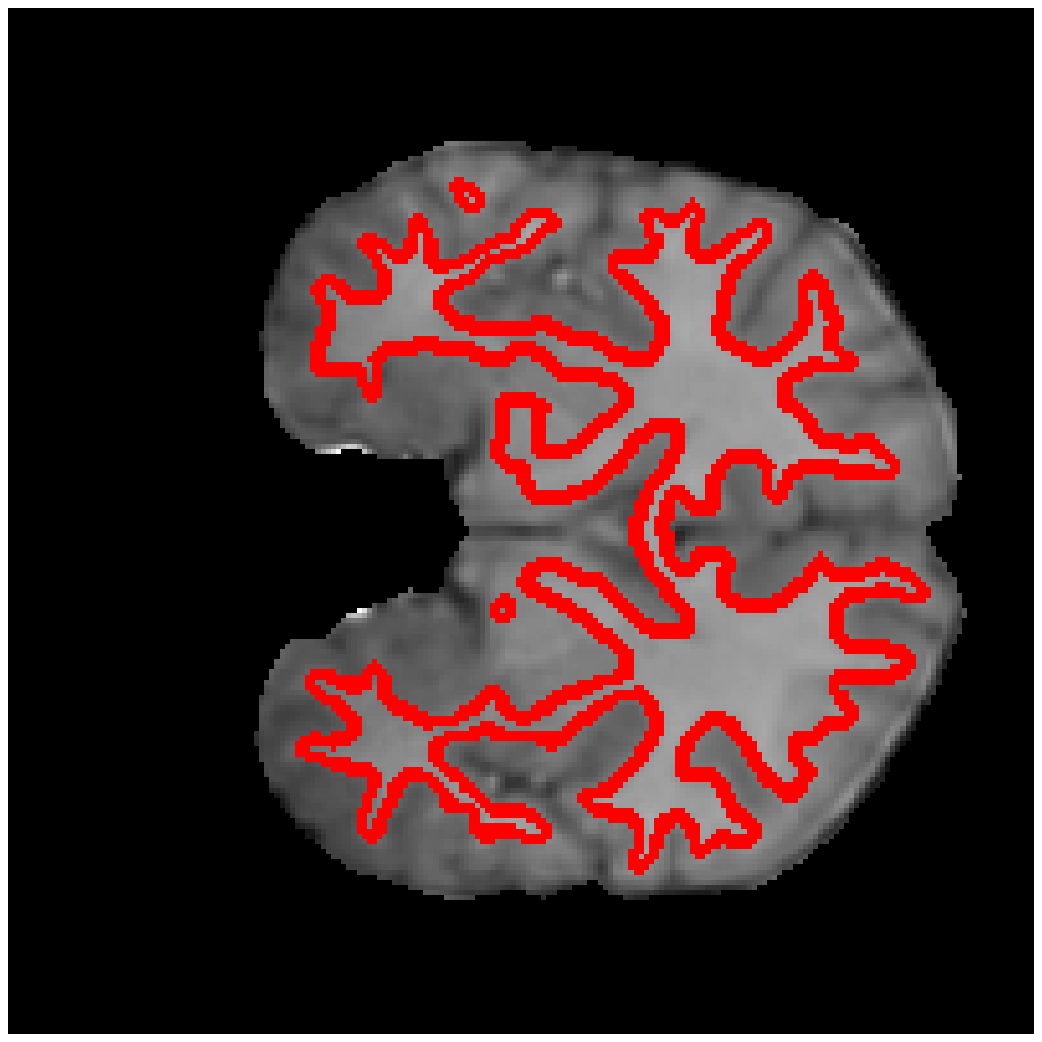}
        
         \hspace{-.5 mm}
         
        \includegraphics[width=0.32\linewidth,height=0.32\linewidth]{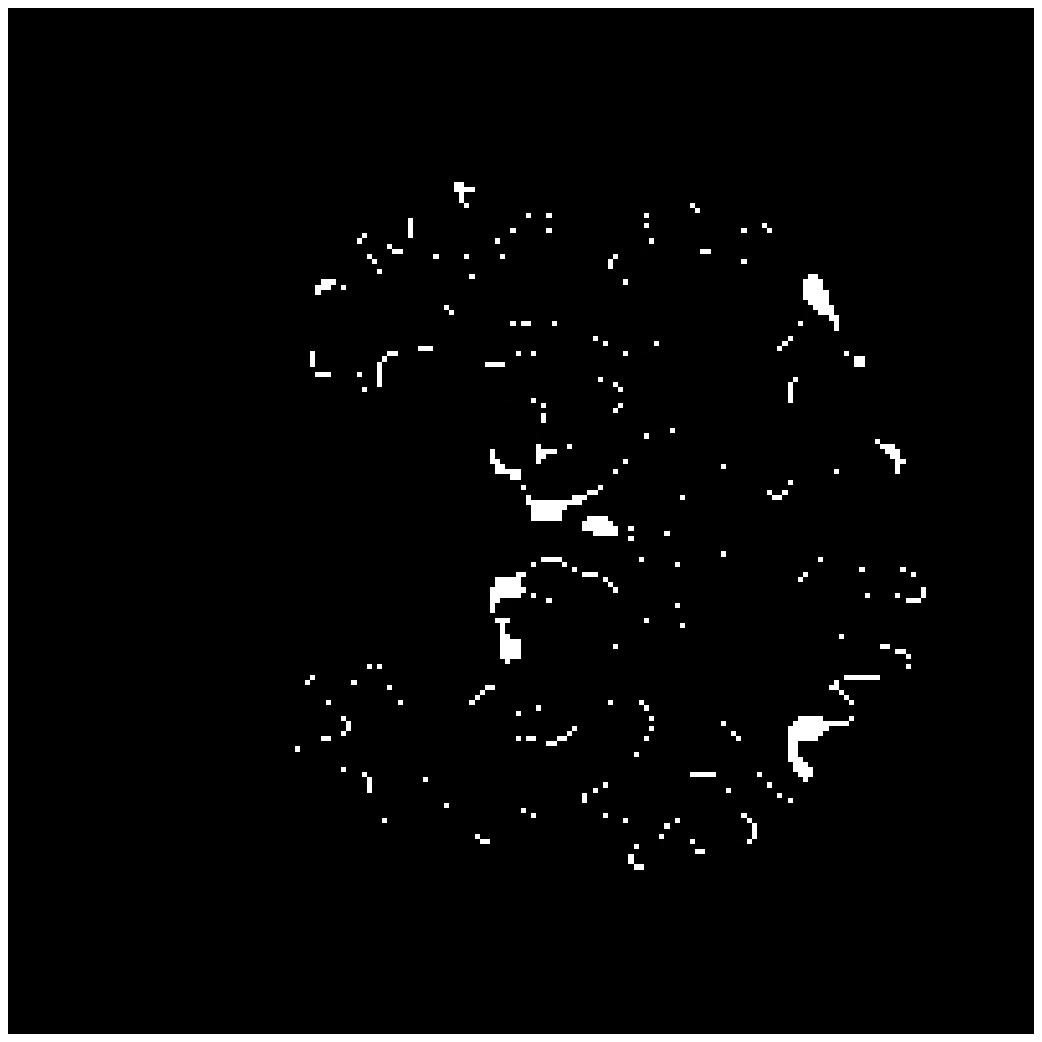}
        }
     \mbox{
      \shortstack{
        \includegraphics[width=0.32\linewidth,height=0.32\linewidth]{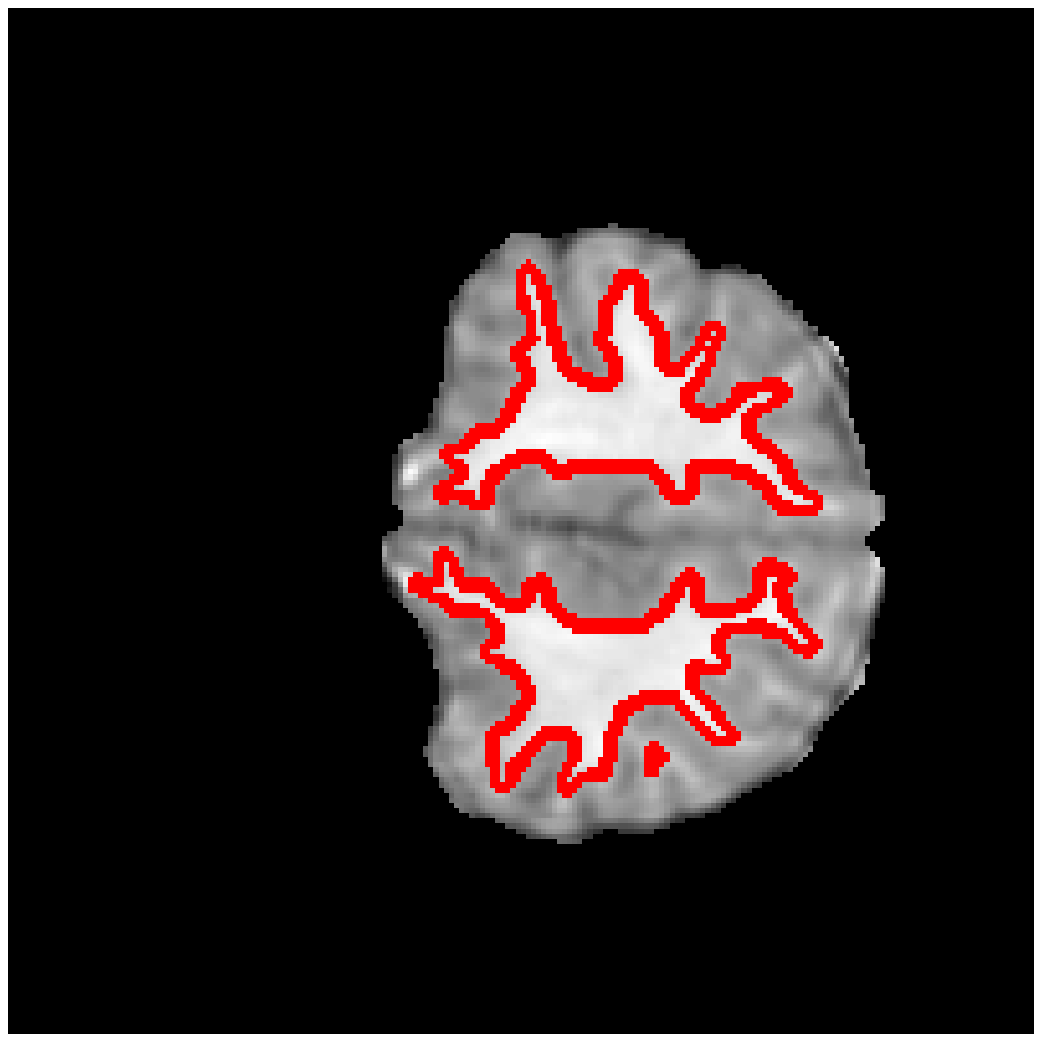} \\
        \dope{} 
        }
        
         \hspace{-.5 mm}
         
      \shortstack{     
        \includegraphics[width=0.32\linewidth,height=0.32\linewidth]{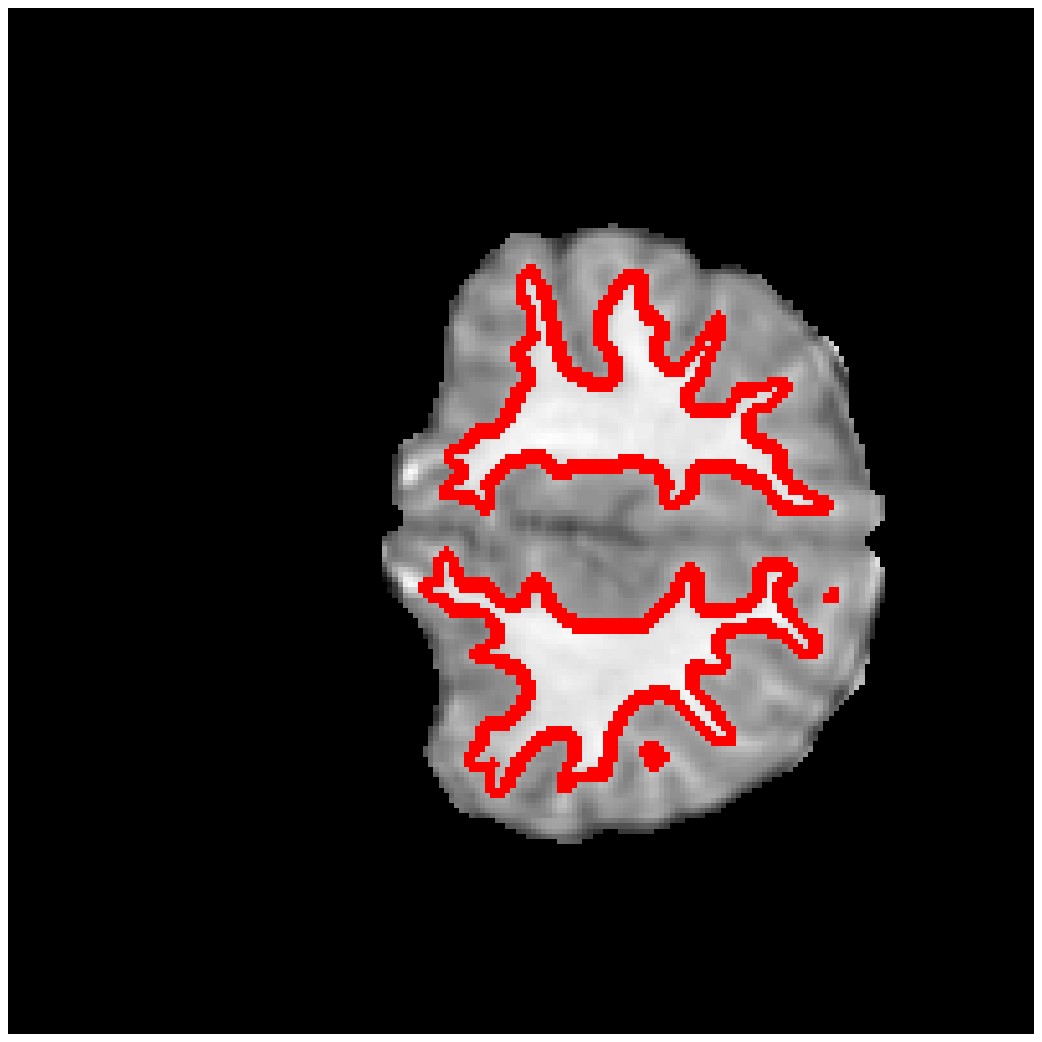} \\
        Graph Cuts
        }
        
         \hspace{-.5 mm}
         
      \shortstack{
        \includegraphics[width=0.32\linewidth,height=0.32\linewidth]{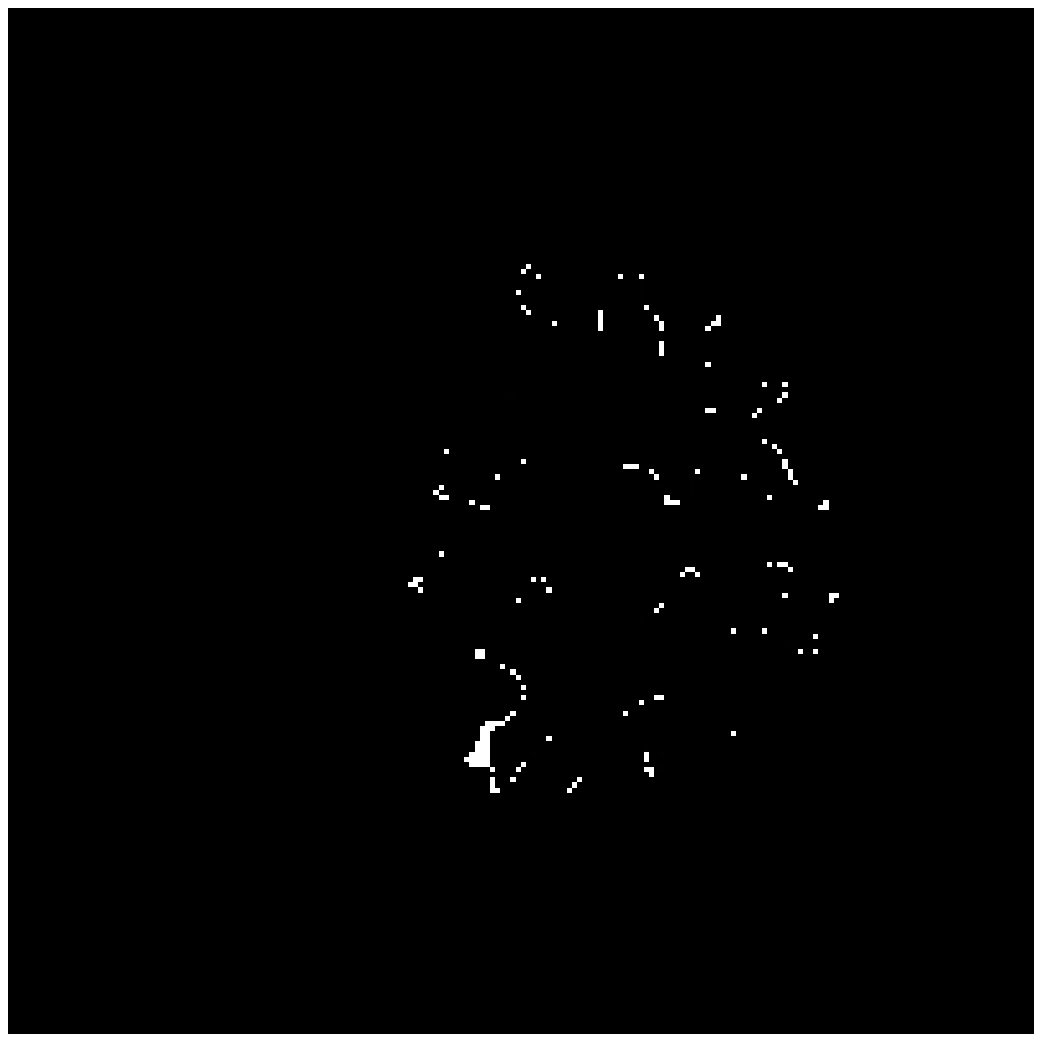} \\
        Difference}
        }
        \caption{Several 2D images in axial view of the 3D segmentation of white matter generated by our \dope{} approach (\emph{left}) and serial graph cuts (\emph{middle}). Differences between both segmentations are shown in the last column.}
        \label{fig:visual3D}
\end{center}        
\end{figure}

\begin{figure}[h!]
     \begin{center}
     \mbox{
        \includegraphics[width=0.48\linewidth,height=0.33\linewidth]{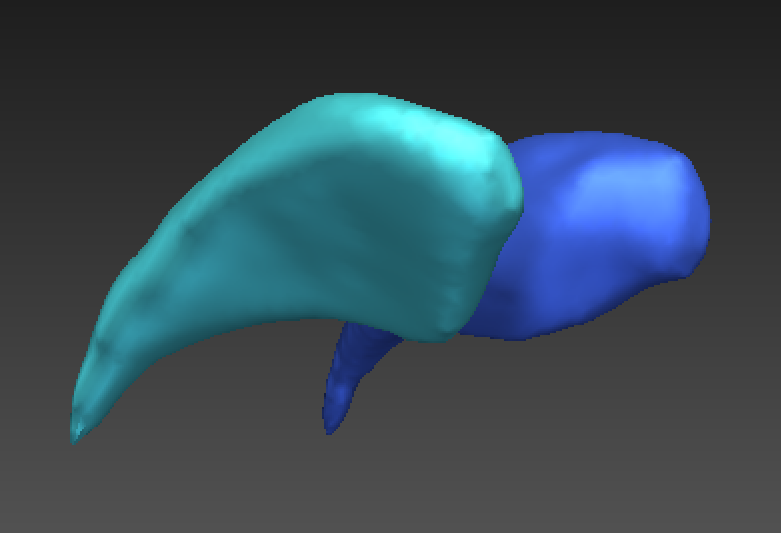}
        
        \hspace{.5mm}
        
        \includegraphics[width=0.48\linewidth,height=0.33\linewidth]{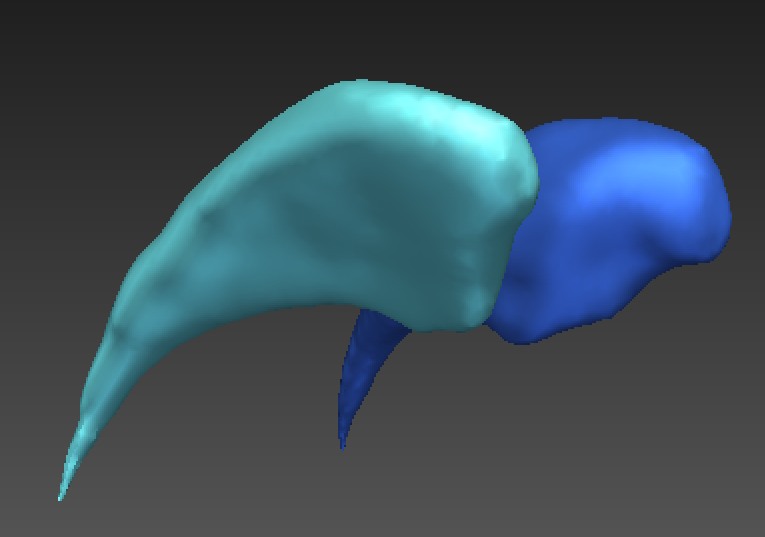}
        }
            
    \caption{3D volumes segmented by serial GC (\emph{left}) and our $\dope$ approach.}
    
   \label{fig:3DVolume}
   \end{center}  
\end{figure}

Figure \ref{fig:visual3D} shows the segmentation of white matter tissues in three axial slices, obtained by our \dope{} method using 128 blocks with $\mr{Size}_{10}$ and kernel size of 7 (\emph{left} column) and sGC (\emph{middle} column). Segmentation differences are shown in the rightmost column. It can be observed that the two segmentations are very similar, with a DSC equal to 0.9638 and an energy difference of only $-1.33\%$. In Figure \ref{fig:3DVolume}, we illustrate 3D segmentation results by showing the surface of the left and right putamen regions, extracted by our method and sGC. Note that, in this case, the probability maps of these specific regions were used as unary potentials in the energy function. 

\subsection{Curvature regularization}
\label{ssection:curv}


We demonstrated that our general formulation can distribute the computations of a powerful serial sub-modular optimization algorithm (i.e., BK) without affecting the quality of the energies at convergence. In this section, we report a curvature regularization experiment to illustrate how our method can also distribute the computations of non-submodular optimization techniques. Specifically, we focused on distributing the computations of the stat-of-the-art LSA-TR method in \cite{Gorelick2014}.  

\begin{figure}[ht!]
 \begin{center}
       \mbox{
      
      \shortstack{
        \includegraphics[width=0.32\linewidth]{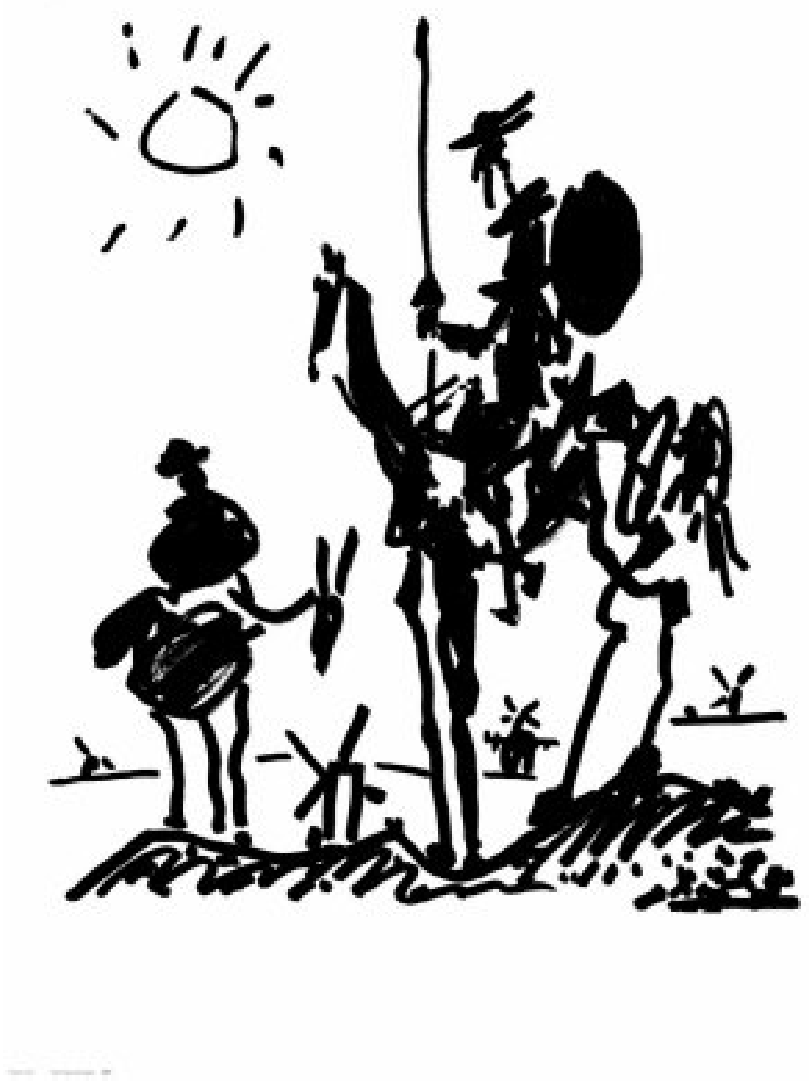} \\
        Original}
        \hspace{-1.75 mm}
         
      \shortstack{
        \includegraphics[width=0.32\linewidth]{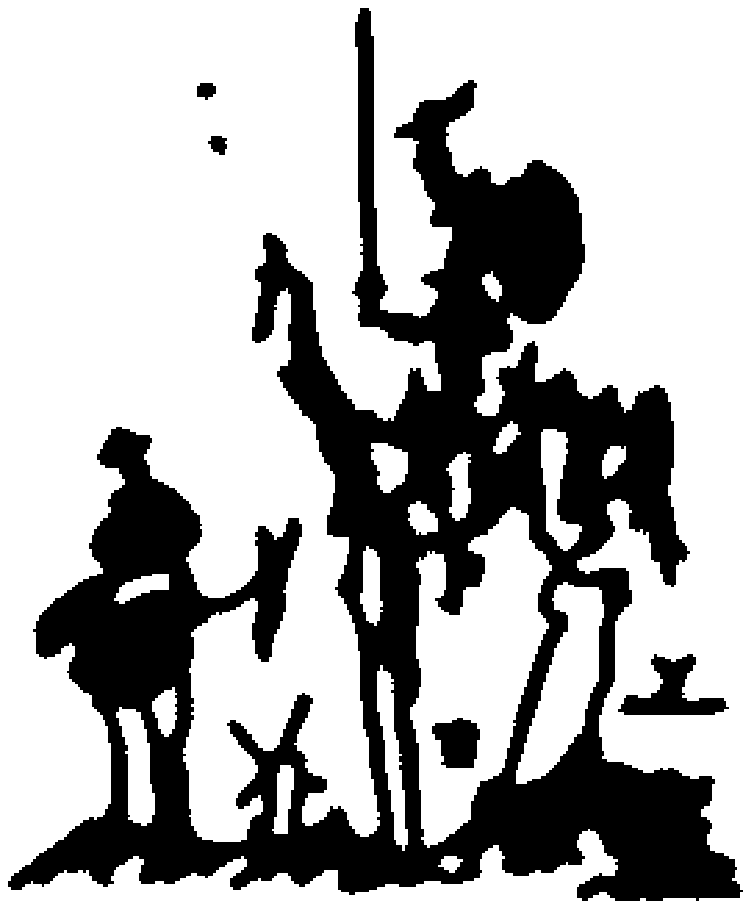} \\
        LSA-TR \cite{Gorelick2014}}
        \hspace{-1.75 mm}
         
      \shortstack{
        \includegraphics[width=0.32\linewidth]{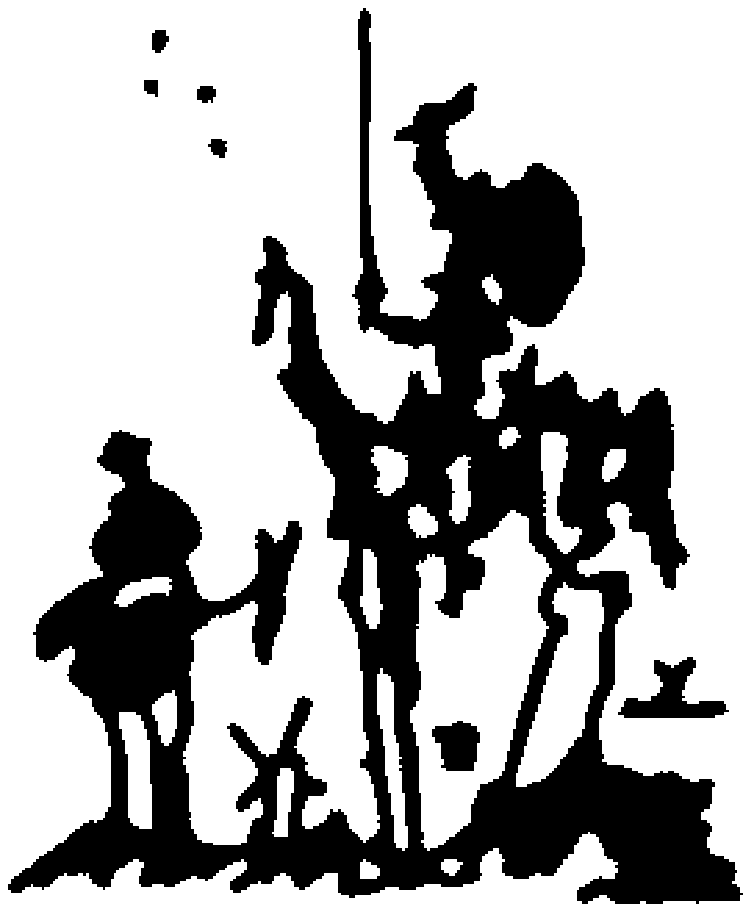} \\
        LSA-TR(\dope{}) }
        
        }

    \caption{Curvature regularization results of Picasso's ink drawing using the trust region (LSA-TR) method in \cite{Gorelick2014} and our \dope{} formulation.}
    
   \label{fig:LSA_TR}
\end{center}
\end{figure}

Table \ref{tab:curvtable} reports the energies obtained by LSA-TR non-submodular optimization \cite{Gorelick2014} and a distributed version based on our \dope{} formulation. To obtain these energies, we employed the squared curvature model and the Picasso's ink drawing used in \cite{Nieuwenhuis2014}. Fig. \ref{fig:LSA_TR} depicts the results of this experiment. Notice that, by distributing the computations of LSA-TR with our \dope{} formulation, we obtained a very similar result while reducing computation time.

\begin{table}[ht!]
\centering
\scriptsize
\begin{tabular}{lcc}
\toprule
\textbf{}             & \multicolumn{1}{l}{\textbf{LSA-TR \cite{Gorelick2014}}} & \multicolumn{1}{l}{\textbf{\begin{tabular}[c]{@{}c@{}} LSA-TR (\dope{}) \\ (8 sub-blocks) \end{tabular}}}  \\ 
\midrule\midrule
\textbf{Energy}  & 1.0906 $\times$ 10$^{4}$  & 1.1115 $\times$ 10$^{4}$                                         \\  \midrule
\textbf{Time}  & 174 s  & 53 s                 \\  
\bottomrule
\end{tabular}
\caption{Energies obtained by LSA-TR non-submodular optimization \cite{Gorelick2014} and our formulation.}
\label{tab:curvtable}
\end{table}

\section{Conclusion}

In this work, we have formulated a general Alternating Direction Method of Multipliers (ADMM) technique that systematically distributes computations in the context of pairwise functions. Our method is scalable, allowing computations for large images by decomposing the problem into a large set of small sub-problems that can be solved in parallel. Unlike existing approaches, which distributes computation to a restricted number of cores \cite{liu2009paint}, our algorithm can be easily adapted to the number of available cores. 
As shown in the results, another advantage of our technique is that it requires a small number of iterations to converge: about 3 iterations for high-definition 2D images and 5-10 iterations for 3D volumes. This allows our method to obtain segmentation results consistent with those of sGC, while allowing distributed computations.

While our experiments have focused on a standard Potts model, one of the main benefits of our formulation lies in its generality. This generality has been proven by distributing the computations in the context of a non-submodular function. Thus, our approach is not technique-specific, and can be applied to a wide variety of pairwise potentials, 
including dense (or fully connected) models. In future work, we plan to extend our method to such models, e.g., the dense pairwise potentials in \cite{Krahenbuhl2011}, and to the case of multi-label segmentation. 

\appendix

\section{Proof of Theorem \ref{th:theo1}}\label{sec:proof}

\expandafter\def\expandafter\small\expandafter{%
    \normalsize
    \setlength\abovedisplayskip{1pt}
    \setlength\belowdisplayskip{3pt}
    \setlength\abovedisplayshortskip{1pt}
    \setlength\belowdisplayshortskip{3pt}
}

\begin{small}

Using relationship \eqref{th:prop1} in Eq. (\ref{eq:cost-function}), and relaxing the integer constraints on $\yy$, general segmentation problem (\ref{eq:cost-function}) can be reformulated as: 
\begin{multline}
\argmin_{\substack{\yy \, \in \, \RR^{|\Om|} \\ \hy_k \, \in \, \{0,1\}^{|\Om_k|} }} \
	\sum_{k=1}^K \tr{\uu} Q^{-1}\tr{S}_k \hy_k \\[-5mm]
		+ \ \lambda \sum_{k=1}^K \sum_{l=1}^K  \tr{\hy}_k S_k Q^{-1} L Q^{-1} \tr{S}_l \hy_l 
\end{multline}
where $\hy_k = S_k \yy$ and $k=1, \ldots, K.$ The unary term can be simplified by defining vectors $\hu_k = S_k Q^{-1} \uu$, and then the problem can be written as:
\vspace{-1mm}
\begin{multline}
	\!\!\!\!\!\!\argmin_{\substack{\yy \, \in \, \RR^{|\Om|} \\ \hy_k \, \in \, \{0,1\}^{|\Om_k|} }}  \
		\sum_{k=1}^K \tr{\hu}_k \hy_k	\ + \ \lambda \sum_{k=1}^K \sum_{l=1}^K  \tr{\hy}_k S_k Q^{-1} L Q^{-1} \tr{S}_l \hy_l.
\end{multline}

We notice that the segmentation vectors of each block are still coupled in the right-most term of this new cost function. In order to segment each block independently, we thus split this term in two: the cost of assigning labels to pixels in the same block and in different blocks:  
\begin{multline}
	\argmin_{\substack{\yy \, \in \, \RR^{|\Om|} \\ \hy_k \, \in  \, \{0,1\}^{|\Om_k|} }} \
		\sum_{k=1}^K \tr{\hu}_k \hy_k
			\ + \ \lambda \sum_{k=1}^K \tr{\hy}_k S_k Q^{-1} L Q^{-1} \tr{S}_k \hy_k \\[-5mm]
				+ \ \lambda \sum_{k=1}^K \sum_{l=1\neq k}^K  \tr{\hy}_k S_k Q^{-1} L Q^{-1} \tr{S}_l \hy_l.
\end{multline}
Using the fact that $\hy_l = S_l \yy$, we can then reformulate the problem as:
\begin{multline}
	\argmin_{\substack{\yy \, \in \, \RR^{|\Om|} \\ \hy_k \, \in  \, \{0,1\}^{|\Om_k|} \\ k=1,\ldots, K}}  \		
		\sum_{k=1}^K \tr{\hu}_k \hy_k \ + \ \lambda \sum_{k=1}^K \tr{\hy}_k S_k Q^{-1} L Q^{-1} \tr{S}_k \hy_k \\[-5mm]
			\ + \ \lambda \sum_{k=1}^K  \tr{\hy}_k S_k Q^{-1} L Q^{-1} \sum_{l\neq k} \tr{S}_l S_l \yy.
\end{multline}
Moreover, since $\sum_{l\neq k} \tr{S}_l S_l = Q - \tr{S}_k S_k$, the problem becomes:
\begin{multline}
	\argmin_{\substack{\yy \, \in \, \RR^{|\Om|} \\ \hy_k \, \in  \, \{0,1\}^{|\Om_k|} }} \	
		\sum_{k=1}^K \tr{\hu}_k \hy_k
			\ + \ \lambda \sum_{k=1}^K \tr{\hy}_k S_k Q^{-1} L Q^{-1} \tr{S}_k \hy_k \nonumber\\[-5mm]
				+ \ \lambda \sum_{k=1}^K  \tr{\hy}_k S_k Q^{-1} L \big(I - Q^{-1} \tr{S}_k S_k\big) \yy.
\end{multline}
Let $C_k = S_k Q^{-1} L (I - Q^{-1} \tr{S}_k S_k)$, we can therefore simplify the cost function as follows:
\begin{multline}
	\argmin_{\substack{\yy \, \in \, \RR^{|\Om|} \\ \hy_k \, \in  \, \{0,1\}^{|\Om_k|} }} \
		\sum_{k=1}^K \tr{\big(\hu_k \, + \, \lambda C_k \yy\big)} \hy_k \\[-5mm]
     		 + \ \lambda \sum_{k=1}^K \tr{\hy}_k S_k Q^{-1} L Q^{-1} \tr{S}_k \hy_k.
\end{multline}
where, again $\hy_k$ has to be equal to $S_k \yy,$ and $k=1, \ldots, K.$ While the blocks are now only coupled via $\yy$, we need to further modify the cost function since the right-most term is not a Laplacian. Using the fact that $L = D - W$, where $D$ is a diagonal matrix such that $d_{ii} = \sum_j w_{ij}$, we obtain that:
\[
  S_k Q^{-1} L Q^{-1} \tr{S}_k \ = \  S_k Q^{-1} D Q^{-1} \tr{S}_k \, - \, S_k Q^{-1} W Q^{-1} \tr{S}_k.
\]
Let $\hW_k = S_k Q^{-1} W Q^{-1} \tr{S}_k$ denote the pairwise potentials of block $k$, adjusted to consider the occurrence of pixels in multiple blocks, and let $\hD_k$ be the diagonal matrix such that  $[\hD_k]_{ii} = \sum_j [\hW_k]_{ij}$. We have:
\begin{multline}
  S_k Q^{-1} L Q^{-1} \tr{S}_k =   S_k Q^{-1} D Q^{-1} \tr{S}_k \, - \, \hW_k \, + \, \hD_k \, - \, \hD_k \\
  \quad  = \big(S_k Q^{-1} D Q^{-1} \tr{S}_k \, - \, \hD_k\big) \, + \, \hL_k. \nonumber
\end{multline}
where $\hL_k$ is the Laplacian of $\hW_k$. Let  $R_k = S_k Q^{-1} D Q^{-1} \tr{S}_k \, - \, \hD_k$, we then use the fact that $\hy_k = S_k \yy$ to reformulate the problem as:
\begin{multline}
	\argmin_{\substack{\yy \, \in \, \RR^{|\Om|} \\ \hy_k \, \in \, \{0,1\}^{|\Om_k|}}} \  
		\sum_{k=1}^K \tr{\big(\hu_k \, + \, \lambda (C_k + R_k S_k) \yy\big)} \hy_k \\[-5mm]	 
			+ \lambda \sum_{k=1}^K \tr{\hy}_k \hL_k \hy_k.
\end{multline}

\end{small}

\bibliographystyle{ieee}
\bibliography{CVPR2017}

\end{document}